\documentclass{article}

\usepackage[ruled,vlined]{algorithm2e}%
\usepackage{arxiv}
\usepackage[square,numbers]{natbib}
\usepackage[utf8]{inputenc} 
\usepackage[T1]{fontenc}    
\usepackage{hyperref}       
\usepackage{url}            
\usepackage{booktabs}       
\usepackage{amsfonts}       
\usepackage{nicefrac}       
\usepackage{microtype}      
\usepackage{lipsum}		
\usepackage{graphicx}
\usepackage{doi}
\usepackage{tabularx}
\usepackage{graphicx}
\usepackage{subcaption}
\usepackage{amsmath}
\usepackage{mdframed}
\usepackage{longtable}
\newcolumntype{P}[1]{>{\raggedright\arraybackslash}p{#1}}
\usepackage{comment}
\usepackage{csquotes}
\usepackage{xcolor}
\usepackage[table]{xcolor}
\usepackage{multirow}
\usepackage{array}
\usepackage{listings}
\usepackage{rotating}   
\usepackage{colortbl}%
\usepackage{threeparttable}   
\usepackage{rotating}
\usepackage{tikz}
\usetikzlibrary{arrows.meta,positioning,shapes.geometric,fit,calc}
\usepackage{pdflscape}
\usepackage{adjustbox}

\usepackage[utf8]{inputenc}
\usepackage{textcomp}
\usepackage{multirow}
\usepackage{amsmath, amssymb}
\usepackage{longtable, multirow}
\usepackage[utf8]{inputenc}%
\newcommand{\good}[1]{\cellcolor{green!20}#1}
\newcommand{\med}[1]{\cellcolor{orange!20}#1}
\newcommand{\bad}[1]{\cellcolor{red!20}#1}

\DeclareUnicodeCharacter{2018}{`}   
\DeclareUnicodeCharacter{2019}{'}   
\DeclareUnicodeCharacter{201C}{``}  
\DeclareUnicodeCharacter{201D}{''}  
\DeclareUnicodeCharacter{2013}{--}  
\DeclareUnicodeCharacter{2014}{---} 
\DeclareUnicodeCharacter{2026}{...} 
\DeclareUnicodeCharacter{2192}{\ensuremath{\to}}     
\DeclareUnicodeCharacter{222A}{\ensuremath{\cup}}    
\DeclareUnicodeCharacter{2248}{\ensuremath{\approx}} 
\DeclareUnicodeCharacter{03A3}{\ensuremath{\Sigma}}  
\DeclareUnicodeCharacter{00D7}{\ensuremath{\times}}  
\DeclareUnicodeCharacter{00B1}{\ensuremath{\pm}}     
\DeclareUnicodeCharacter{2264}{\ensuremath{\le}}     
\DeclareUnicodeCharacter{2212}{-}                    

\usepackage[most]{tcolorbox}
\tcbuselibrary{listings,breakable,skins}
\newtcblisting{PromptBlock}{
  enhanced,
  breakable,
  colback=black!2,
  colframe=black!35,
  boxrule=0.4pt,
  arc=2mm,
  left=2mm,right=2mm,top=1.5mm,bottom=1.5mm,
  listing only,
  listing options={
    basicstyle=\ttfamily\footnotesize,
    breaklines=true,
    columns=fullflexible,
    keepspaces=true,
    showstringspaces=false
  }
}

\title{Forecasting Commencing Enrolments Under Data Sparsity: A Zero-Shot Time Series Foundation Models Framework for Higher Education Planning}

\author{ 
  \href{https://orcid.org/0009-0003-7339-9294}{\includegraphics[scale=0.06]{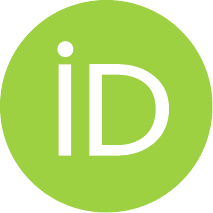}\hspace{1mm}Jittarin Jetwiriyanon}\thanks{Corresponding author: jittarin.jetwiriyanon.1@uni.massey.ac.nz} \\ 
  School of Mathematical and Computational Sciences\\%
  Massey University\\%
  Albany, New Zealand
  \and
  \href{https://orcid.org/0000-0001-9416-1435}{\includegraphics[scale=0.06]{orcid.pdf}\hspace{1mm}Teo Susnjak}\thanks{Contributing author: T.Susnjak@massey.ac.nz}\\%
  School of Mathematical and Computational Sciences\\%
  Massey University\\%
  Albany, New Zealand
  \and
  \href{https://orcid.org/0000-0003-0701-0204}{\includegraphics[scale=0.06]{orcid.pdf}\hspace{1mm}Surangika Ranathunga}\thanks{Contributing author: S.Ranathunga@massey.ac.nz}\\%
  School of Mathematical and Computational Sciences\\%
  Massey University\\%
  Albany, New Zealand
}

\renewcommand{\shorttitle}{Zero-Shot Time Series Foundation Models for Annual Institutional Forecasting Under Data Sparsity}

\begin{document}
\maketitle

\begin{abstract}
Effective resource allocation in higher education depends on reliable enrolment forecasts, yet institutional planners frequently face data series disrupted by structural shifts. This paper investigates whether zero-shot Time Series Foundation Models (TSFMs) can provide rigorous decision support for annual enrolment forecasting under severe data sparsity. We benchmark multiple TSFMs against classical operational baselines using an expanding-window backtest that mirrors decision-time constraints. To capture environmental shifts without look-ahead bias, we introduce a leakage-safe covariate protocol that integrates feature-engineered Google Trends with the Institutional Operating Conditions Index (IOCI), a transferable regime measure extracted from historical narrative evidence. Our evaluation demonstrates that covariate-conditioned TSFMs are competitive with classical methods and can improve accuracy without requiring bespoke institutional training. However, the operational benefits depend on cohort characteristics and covariate design. This study provides an auditable and transferable forecasting protocol for operational researchers and university administrators, helping institutions determine when context-aware forecasting adds practical value under limited data and structural instability.
\end{abstract}


\section{Introduction}
\label{introduction}
Accurate forecasting of annual new-student enrolments is central to strategic and operational planning in higher education, directly driving staffing, teaching capacity, and workload optimisation \citep{KUO20096108, HUANG20118014}. The need for reliable models has grown as universities face tightening budgets, economic volatility, and demographic shifts like the US enrollment cliff \citep{kelchen2024, grawe2018, bird2023, willis2024}. Under these pressures, static historical averaging and ratio-based projections frequently fail at turning points \citep{Long2013CrisisEnrollment}. Consequently, institutions require dynamic, data-driven forecasting frameworks to manage deficits, optimise resource allocation, and ensure long-term sustainability \citep{Sharkey01022025, mendez2025, oecd_fin_sust_2025, oecd_eag_2024}.

Institutional enrolment series present significant forecasting challenges because they are short, highly aggregated, and restricted by privacy concerns \citep{MALTZ2007106, LI2024122203}. Even with multi-year histories, the effective number of training observations shrinks once structural breaks and cohort reporting changes are considered. In such low signal-to-noise environments, persistence outperforms richer models, which tend to overfit unless anchored by credible exogenous structure \citep{MAKRIDAKIS202054}. This motivates the development of context-aware, multivariate forecasting approaches that incorporate external drivers while maintaining operational temporal discipline \citep{MALTZ2007106, BERGMEIR201870}.

Recent TSFMs provide a practical option for these data-sparse environments \citep{BOLANOSMARTINEZ2025126245, ansari2024chronos, das2024decoderonlyfoundationmodeltimeseries, woo2024moirai}. Pre-trained on vast heterogeneous datasets, TSFMs provide strong zero-shot priors and allow conditioning on dynamic covariates without requiring expensive, institution-specific fine-tuning \citep{woo2024moirai,das2024decoderonlyfoundationmodeltimeseries}. Furthermore, operational research currently lacks practical guidance on constructing leakage-safe covariates under extreme sparsity and identifying when these signals genuinely improve model stability \citep{BERGMEIR201870}.

Public web signals, such as search-intensity measures and unstructured text, provide a pragmatic source of external structure to proxy latent demand and broader environmental shifts \citep{ChoiVarian2012}. To prevent data leakage and spurious fit in data-sparse settings \citep{GRUBER20251589}, these signals can be engineered with strict temporal discipline \citep{BERGMEIR201870}. Because search-platform indices are dynamically normalised, operational practice requires explicit vintage control, aggregation to the decision frequency, and stabilising transformations \citep{EichenauerEtAl2022ConsistentGT}. By constructing covariates exclusively at decision time, they safely inject exogenous information without contaminating the forecast origin.

This same discipline applies to unstructured narrative sources, which can be treated as quantitative evidence \citep{grimmer2013_text_as_data}. Operationally, this involves a transparent coding scheme that maps time-stamped text into numerical features like sentiment, before aggregating them to the decision frequency. This mirrors established text-derived uncertainty indices that translate qualitative narratives into interpretable time series \citep{BakerBloomDavis2016EPU}. 

This study addresses that gap by developing and evaluating a decision-time forecasting framework for annual commencing-enrolment planning under severe data sparsity. The framework combines an expanding-window backtest with leakage-safe covariate construction and compares zero-shot TSFMs against operational benchmark models. External information is introduced through feature-engineered Google Trends demand proxies and IOCI, a year-specific contextual covariate derived from contemporaneous narrative evidence. The emphasis is not only on forecast accuracy, but also on auditability, calibration, and the conditions under which context-aware forecasting is operationally worth deploying.

\subsection*{Contribution}
This paper contributes to operational research in institutional planning by developing and evaluating a forecasting workflow for annual commencing-enrolment decisions under severe data sparsity. The study frames annual commencing-enrolment forecasting as a decision-time planning problem in which historical series are short, structural change is common, and external information can be incorporated without look-ahead bias. Within this setting, we benchmark zero-shot TSFMs against interpretable operational baselines and examine when leakage-safe contextual covariates improve performance. The contribution is a transferable forecasting protocol for data-sparse planning environments, supported by an auditable covariate construction procedure and evidence on model-covariate compatibility under annual operational constraints.

\begin{itemize}
\item \textbf{A decision-time forecasting framework for annual planning:} We develop an expanding-window evaluation protocol that reflects institutional planning practice by enforcing strict forecast-origin discipline and leakage-safe preprocessing.
\item \textbf{An operational comparison of forecasting designs under data sparsity:} We compare zero-shot TSFMs with strong benchmark models to identify which forecasting designs deliver the most reliable accuracy–robustness trade-off across domestic and international enrolment cohorts.
\item \textbf{An auditable external-information protocol:} We show how public web signals and narrative institutional evidence can be converted into operational covariates through transparent engineering rules that preserve temporal validity and reproducibility.
\item \textbf{Guidance on when contextual forecasting is worth the added complexity:} We demonstrate that external covariates can materially improve forecasts in some configurations, but that their value depends on the interaction between cohort characteristics, covariate stability, and model architecture.
\end{itemize}

\section{Related Work}\label{sec2}
\subsection{Enrolment forecasting under institutional constraints}
Early operational planning for enrolments relied predominantly on univariate heuristics, such as grade-progression ratios and cohort-survival calculations, to guide capacity and resource allocation in schools and colleges \citep{Webster01091970}. While persistence (naïve) baseline remains effective over very short horizons due to its computational simplicity \citep{SBRANA2023114103}, modern institutional planning increasingly demands multivariate statistical models that can capture complex dynamics. State-space formulations have been widely adopted to integrate exogenous predictors into decision-support systems without requiring planners to rebuild models from scratch \citep{HO1998213, Chen_Li_Hagedorn_2019}.

This evolution aligns with broader macro-forecasting practices in operational research, where factor models and data fusion techniques are heavily utilised to mitigate the risks associated with low signal-to-noise environments \citep{Stock01042002, 10.1093/jrsssa/qnae094}. However, forecasting annual institutional demand remains notoriously difficult because these administrative series are short, highly aggregated, and frequently disrupted by regime shifts \citep{caserta2025intermittent}. In such sparse data environments, traditional models lose strength, highlighting the need for dynamic, data-driven frameworks capable of managing deficits and optimising resource allocation when historical patterns are broken by fundamental surprises \citep{Sharkey01022025, mendez2025}.

\subsection{Covariates in student enrolments}
Strategic capacity planning benefits from forecasting models that capture key exogenous drivers of student demand \citep{XIE2023119652}. Traditional economic and demographic indicators, such as high school graduation rates, provide a grounded, baseline expectation for long-term capacity planning \citep{BarrTurner2015}. However, relying solely on lagging economic indicators is insufficient for proactive decision-making. 

Recently, digital search trends, mobility data, and web-scraped public signals have emerged as leading indicators, offering planners early warnings of behavioural shifts before they materialise in administrative enrolment records \citep{ChoiVarian2012, AskitasZimmermann2009, BakerBloomDavis2016EPU, Shapiro2022, Ilin2021}. Integrating these diverse, high-frequency signals transitions institutional forecasting from static trend extrapolation to dynamic, condition-responsive decision support \citep{Slim2018EDM}. In environments characterised by structural breaks, operational forecasting requires the disciplined integration of these exogenous variables and contextual expert information to stabilise baseline models \citep{nikolopoulos2019relative}.

\subsection{Feature engineering for covariates under data sparsity}
Recent operational research highlights that in low signal-to-noise environments, predictive accuracy can be significantly improved by anchoring models with external priors and context-aware structure \citep{Wang01022025}. However, despite advances in model architectures, feature engineering remains the critical step for translating raw, noisy data into operational signals \citep{LimArik2021}.

To prevent data leakage and noise fit, simple transformations such as temporal lags are used to inject short-run persistence into models while strictly preserving the decision-time boundary. Furthermore, window summary statistics like moving averages and exponential smoothing are standard operational techniques for isolating underlying medium-term trends from period-to-period volatility \citep{Brown1959}. For modern predictive architectures, the careful engineering, selection, and vintage-alignment of these time-varying covariates are essential to ensure models remain stable, auditable, and practically deployable in a live institutional setting \citep{Christ2018tsfresh, BERGMEIR201870}.

\subsection{Large language models for institutional narratives and stress signals}
Large Language Models (LLMs) such as recent GPT, Gemini, and Claude variants now offer operational researchers scalable tools to extract structured, time-indexed indicators of institutional stress from unstructured narrative text \citep{Halterman_2025, WU2026129483}. A growing body of evaluation literature shows that these systems can match, and sometimes exceed, the reliability of traditional human coding, making rubric-based annotation feasible even when task-specific, labelled numerical data is scarce \citep{mellon2024_mii_coding_llms}. 

In macroeconomic planning, similar text-as-data techniques are used to convert central bank communications and policy documents into crisis-sensitive risk indices \citep{silva2025_central_bank_comm_llm}. For higher education planning, this approach provides a transparent, decision-aligned methodology to quantify operating conditions into a leakage-safe covariate. Provided that retrospective verification and stability checks are rigorously documented, this methodology effectively operationalises narrative evidence into exogenous predictors that can be safely fed into forecasting models \citep{Atreja_2025, mirandabelmonte2023epu, grimmer2013_text_as_data}.

\subsection{Zero-shot TSFMs with covariates in student enrolment forecasting}
\begin{figure}[ht!]
  \centering
  \includegraphics[width=1.0\textwidth]{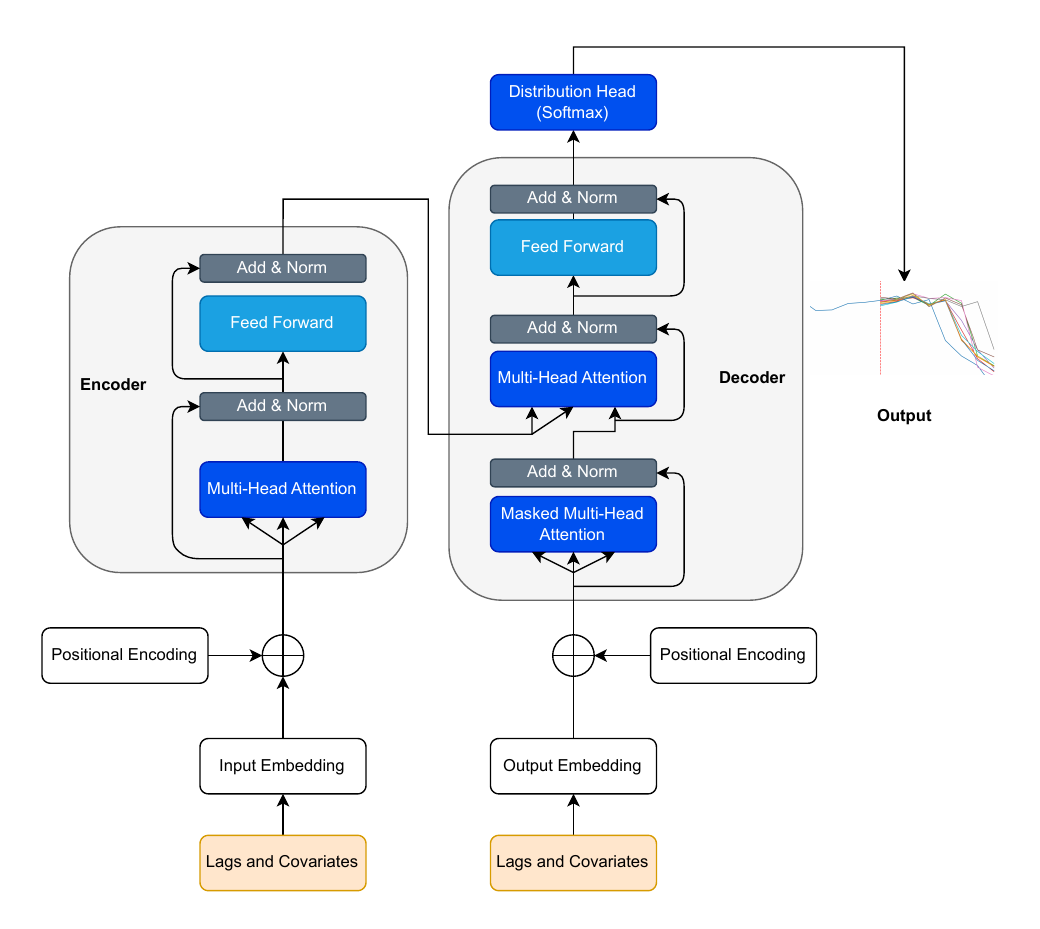}
  \caption{TSFMs architecture for multivariate time series forecasting \citep{math13050814}.}
  \label{fig:transformer_covariates}
\end{figure}

A major development in time series forecasting is the rise of TSFMs, large pre-trained models designed to serve as general-purpose forecasters across domains \citep{kottapalli2025foundationmodelstimeseries} (Figure~\ref{fig:transformer_covariates}). These models, inspired by the success of LLMs, are trained on massive collections of time series data and can perform forecasting in a zero-shot manner \citep{jetwiriyanon2025generalisationboundszeroshoteconomic, liu2024generaltimetransformer}. Several notable TSFMs have emerged, Moirai from Salesforce \citep{woo2024moirai, aksu2024gifteval}, TimesFM from Google \citep{das2024decoderonlyfoundationmodeltimeseries, Vishwas2025}, and Chronos-Bolt from AWS \citep{ansari2024chronos, ansari2025chronos2, wolff2025usingpretrainedllmsmultivariate}. Many TSFMs can incorporate covariates by receiving them as additional input channels alongside the target history. In masked-encoder architectures, dynamic covariates can be provided as parallel variates within the input sequence, enabling the attention mechanism to condition forecasts on external signals \citep{gao2024unitsunifiedmultitasktime, ansari2024chronos}. This capability is attractive for student enrolment forecasting, where institutional histories are short, and privacy constraints limit task-specific training. The convergence of TSFMs and domain-specific covariates offers a promising path for student enrolment forecasting in the common scenario of scarce institutional data.

\clearpage

\subsection{Summary and research questions}
Although TSFMs have shown strong performance on public forecasting benchmarks, there is limited evidence on their value in institutional planning environments characterised by short annual histories, structural instability, and strict decision-time information constraints. In these settings, the practical question is not simply which model achieves the lowest average error, but which forecasting design offers the most reliable and auditable support for planning decisions. This study evaluates forecasting performance as an operational design problem, examining not only zero-shot model accuracy but also the role of external information, covariate stability, and probabilistic reliability under severe data sparsity.

\begin{itemize}
\item \textbf{RQ1:} Which forecasting design provides the most reliable accuracy-robustness trade-off for annual institutional planning under severe data sparsity?
\item \textbf{RQ2:} Under what conditions do leakage-safe external indicators add sufficient practical value to justify their use in operational enrolment forecasting?
\item \textbf{RQ3:} Which covariate construction strategies are sufficiently auditable and transferable for data-sparse institutional forecasting workflows?
\end{itemize}

\section{Methodology}\label{sec3}
\subsection{Dataset}
We evaluated the proposed forecasting pipeline using annual commencing student headcounts (new enrolments) extracted from a public university's administrative system in a data-sparse institutional planning setting, disaggregated into domestic and international cohorts. Although the study was conducted at a single university, the proposed workflow is transferable to forecasting tasks in other data-sparse institutional settings. The outcome variable was defined as the administrative headcount of new entrants, consistent with institutional reporting practice. Pre-processing was intentionally minimal. We harmonised cohort definitions across years and applied adjustments to ensure consistency. Where historical reporting introduced minor reconciliation discrepancies, we allowed a small numerical tolerance in validation checks while retaining the original measurement scale. The dataset comprised two annual target series, namely domestic and international commencing-enrolment headcounts spanning 2007-2025 (\(19\) annual observations per series), which served as the target variables in the evaluation.

\subsubsection{Feature engineering for Google Trends}
Google Trends web signals were extracted using a consistent query term representing the institution's name across worldwide and domestic search geographies. We followed a standardised extraction procedure in which the institution's name served as the query anchor, allowing consistent replication across different entities \citep{medeiros2021properusegoogletrends, KOHNS20231384}. In this study, we used Google Trends Relative Search Volume (RSV) from the worldwide query. RSV is scaled from \(0\) to \(100\) within each extraction window and reflects relative popularity. We aggregated monthly RSV into annual summaries aligned with the enrolment decision frequency, standardised the features within each training window, and constructed leakage-safe covariates commonly used as demand proxies \citep{djorno2025restoringforecastingpowergoogle}.

\paragraph{Feature engineering for covariates}
Engineered summaries were used to improve stability when covariates were supplied to the forecasting models. Let \(\{x_t\}_{t=1}^{T}\) denote the annual scalar input series used to construct a covariate, indexed by year \(t\). Let \(m_t^{(s)}\) denote Exponentially Weighted Moving Average (EWMA) of \(x_t\) with span parameter \(s\) in years and smoothing factor \(\alpha_s \in (0,1]\). We used the common span parameterisation \(\alpha_s = \tfrac{2}{s+1}\) and initialised the recursion with \(m_1^{(s)} = x_1\).

\paragraph{Exponentially weighted moving average (2-year)}
The \(2\)-year EWMA smoothed short-run noise while retaining recent movements. Using the common span parameterisation \(\alpha_s = \tfrac{2}{s+1}\), we set \(\alpha_2 = \tfrac{2}{3}\) and defined the recursion in Equation~\ref{eq:ewma2}.
\begin{equation}
\begin{aligned}
m_t^{(2)} &= \alpha_2 x_t + (1-\alpha_2)m_{t-1}^{(2)},
\qquad m_1^{(2)} = x_1
\end{aligned}
\label{eq:ewma2}
\end{equation}
Compared with a simple moving average, the EWMA places greater weight on recent observations, allowing it to adapt more quickly to level shifts.

\paragraph{Exponentially weighted moving average (3-year)}
A slightly smoother variant emphasised the medium-run signal with lower volatility. With \(\alpha_3 = \tfrac{2}{4} = \tfrac{1}{2}\), we defined Equation~\ref{eq:ewma3}.
\begin{equation}
\begin{aligned}
m_t^{(3)} &= \alpha_3 x_t + (1-\alpha_3)m_{t-1}^{(3)},
\qquad m_1^{(3)} = x_1
\end{aligned}
\label{eq:ewma3}
\end{equation}
Relative to the \(2\)-year EWMA, this covariate was less reactive but more robust to noise.

\paragraph{Lag-1 level}
To capture short-run persistence as an exogenous predictor, we included the lag-\(1\) level, as defined in Equation~\ref{eq:lag1}.
\begin{equation}
\begin{aligned}
\ell_t^{(1)} &= x_{t-1}, \qquad t \ge 2.
\end{aligned}
\label{eq:lag1}
\end{equation}

\subsubsection{Operationalisation of institutional operating conditions}
We specified a reproducible procedure for constructing the IOCI as a template for converting time-stamped institutional narrative evidence into a comparable regime covariate. Using a fixed prompt schema and scoring rubric (\ref{secA1}), LLMs mapped the year-\(t\) evidence pack to a consistent \(0\)-\(100\) measure of operating pressure, yielding an annual scalar \(\text{IOCI}_t \in [0,100]\). Where necessary for interpretation, the model could consult reputable public context explicitly dated to the scored year, but the score and justification remained anchored to the supplied institutional evidence. Unlike economy-wide uncertainty measures, the IOCI was designed as an institution-level, decision-time covariate for enrolment forecasting. Evidence was restricted to information available for year \(t\), and explicit vintage constraints were enforced to prevent look-ahead bias.

Although the IOCI evidence referenced international enrolment fee dependency, the construct was designed to reflect institution-wide operating conditions. In many university systems, international fee income cross-subsidises domestic teaching and broader institutional activity, so shocks to international income are transmitted directly into domestic-facing operations. This makes the signal plausibly predictive of domestic enrolment outcomes as well.

To make the procedure practical and repeatable, we wrote a dedicated system prompt that specified the analyst role, the scoring scale, and explicit constraints on evidence use. The prompt framed (\ref{secA2}) the model as a university planning, administrative analyst and instructed it to rate operating stress on a \(0\)-\(100\) scale based only on the supplied text. We also built guardrails into the process, including year-specific scoring, the avoidance of assumptions when evidence was unclear, and explicit rationales linking evidence to scores. In this way, the process remained transparent and replicable, and the prompt itself constituted a reusable contribution for other institutional settings.

Table~\ref{tab:model_scores_calibrated} shows a retrospective verification of the IOCI construction procedure using four large language model (LLM) configurations, Grok Fast 1, Gemini 3.1 Pro, Claude Opus 4.6, and GPT-5.4 Thinking. Specifically, we applied the fixed prompt to the historical, year-indexed evidence and compared the resulting model-derived IOCI scores with an existing reference series. Concordance between the two series provides supporting evidence that the prompt operationalises the intended construct and yields a reproducible mapping from narrative evidence to a quantitative regime covariate. The calibrated IOCI series treats the GPT-5.4 Thinking outputs as the primary signal and applies a bounded post-hoc adjustment of \(\pm 1\) point to improve level alignment. This calibration is intentionally conservative: it preserves the year-to-year ordering while correcting minor systematic offsets and reducing small inconsistencies attributable to stochastic variation in model judgements. Thus, IOCI is a methodology for operationalising qualitative text as a time-series covariate. This technique, which uses an LLM to score historical narratives without look-ahead bias, can be applied to any domain in which numerical data are sparse.

\begin{table}[htbp]
\centering
\caption{Model-generated series by year.}
\label{tab:model_scores_calibrated}
\begin{tabular}{cccccc}
\toprule
\textbf{} & \textbf{Grok} & \textbf{Gemini 3.1} & \textbf{Claude} & \textbf{GPT 5.4} & \textbf{} \\
\textbf{Year} & \textbf{Fast 1} & \textbf{Pro} & \textbf{Opus 4.6} & \textbf{Thinking} & \textbf{Calibrated} \\

\midrule
2007 & 8  & 10 & 12 & 15 & 15 \\
2008 & 8  & 10 & 12 & 15 & 15 \\
2009 & 8  & 10 & 12 & 15 & 15 \\
2010 & 17 & 19 & 17 & 20 & 21 \\
2011 & 13 & 14 & 14 & 6  & 7  \\
2012 & 14 & 15 & 16 & 7  & 6  \\
2013 & 14 & 15 & 16 & 7  & 7  \\
2014 & 53 & 46 & 48 & 48 & 49 \\
2015 & 55 & 49 & 50 & 51 & 51 \\
2016 & 57 & 50 & 52 & 53 & 54 \\
2017 & 29 & 35 & 38 & 39 & 39 \\
2018 & 59 & 47 & 50 & 50 & 51 \\
2019 & 57 & 55 & 57 & 57 & 58 \\
2020 & 95 & 86 & 92 & 85 & 86 \\
2021 & 94 & 90 & 95 & 95 & 95 \\
2022 & 92 & 88 & 90 & 94 & 94 \\
2023 & 71 & 75 & 73 & 74 & 75 \\
2024 & 50 & 58 & 55 & 58 & 59 \\
2025 & 30 & 38 & 35 & 39 & 39 \\
\bottomrule
\end{tabular}
\end{table}
\clearpage

\subsection{Zero-shot forecasting with covariates}
We used zero-shot TSFMs to generate enrolment forecasts while keeping all model weights fixed. In this setting, the pre-trained model served as a strong domain prior learned from large and heterogeneous collections of time series. This strategy was well-suited to institutional contexts with short histories and privacy constraints because it avoided dataset-specific training while enabling the use of external signals without exposing individual-level records.

Let \(y_t\) denote annual new-enrolment counts for each cohort. We assembled a covariate matrix \(\mathbf{X}_t\) comprising internal administrative indicators, public attention proxies, and structural features available at the forecast origin. For each forecast origin \(t\) in an expanding-window backtest, we passed a multivariate context of length \(L\), as shown in Equation~\ref{eq:covariates}:
\begin{equation}
\mathcal{C}_t = \big\{(y_{t-L+1},\mathbf{X}_{t-L+1}),\ldots,(y_t,\mathbf{X}_t)\big\}.
\label{eq:covariates}
\end{equation}
For TSFMs that supported dynamic covariates over the forecast horizon, we supplied covariate history only up to time \(t\); no future values were injected into the reported results. When scenario analysis was of interest, future covariate paths could be supplied to obtain counterfactual forecasts.

\subsection{Baseline models}
We included interpretable reference models to quantify the incremental value of the proposed forecasting experiments. These baselines clarified the level of performance that was achievable under a stringent evaluation setting.

\subsubsection{Persistence}
The persistence benchmark projected the last observed value forward across all forecast horizons by repeating the most recently observed level over the entire forecast window \citep{MakridakisHibon2000}. Despite its simplicity, persistence was competitive in short samples and served as a conservative lower bound for acceptable performance. Evidence from large-scale forecasting competitions has shown that simple benchmarks can rival more complex methods, especially in the presence of structural change and limited data \citep{MAKRIDAKIS202054}. Persistence is appropriate for series that are well approximated by a random walk, and more elaborate techniques cannot systematically outperform the na\"ive forecast \citep{Hamilton1994}. Recent applications have also documented the practicality of persistence-style baselines in operational settings \citep{SBRANA2023114103}. Consequently, persistence provided a realistic baseline for assessing whether more complex models delivered genuine predictive value.

Given a series \(\{y_t\}_{t=1}^{T}\) and forecast horizon \(H\), the persistence (na\"ive) forecast carried the last observed level forward, as defined in Equation~\ref{eq:persistence}:
\begin{equation}
\widehat{y}_{t+h \mid t} = y_t, \qquad h = 1,\dots,H.
\label{eq:persistence}
\end{equation}

\subsubsection{ARIMA/ARIMAX}
We adopted a non-seasonal Autoregressive Integrated Moving Average (ARIMA) as a classical baseline. When external covariates were included, we used the corresponding Autoregressive Integrated Moving Average with Exogenous variables (ARIMAX) specification \citep{BoxJenkins2015}. This model comprises autoregressive terms, differencing, moving-average terms, and exogenous regressors \citep{Pankratz1991}. All parameters were re-estimated at each expanding forecast origin to adapt to evolving dynamics over time \citep{Tashman2000}. Model adequacy was assessed using residual diagnostics, including checks of residual whiteness, to ensure that comparisons against TSFMs were based on well-specified classical baselines \citep{LjungBox1978}, parameters for the non-seasonal ARIMA/ARIMAX$(p,d,q)$.
\begin{itemize}
    \item \(p\): the number of past values of the series included in the autoregressive component
    \item \(d\): the degree of differencing required to make the series more nearly stationary
    \item \(q\): the number of past error terms included in the moving-average component
    \item exogenous regressors: external variables added to improve forecasting performance
\end{itemize}

\subsection{Time series foundation models}
We benchmarked three pre-trained TSFM families in zero-shot settings with covariates. All models received the student enrolment history and, when applicable, a matrix of known covariates. Inputs were standardised using parameters computed from the training window and then transformed back to the original scale at the output stage.

\subsubsection{Moirai}
Moirai-Small comprises a 6-layer masked encoder with model width \(d_{\text{model}} = 384\), feed-forward width \(d_{\text{ff}} = 1536\), and \(n_{\text{heads}} = 6\) attention heads, for approximately 14M parameters. Moirai-Base increases depth and width to 12 layers with \(d_{\text{model}} = 768\), \(d_{\text{ff}} = 3072\), and \(n_{\text{heads}} = 12\), for approximately 91M parameters. Moirai-Large scales to 24 layers with \(d_{\text{model}} = 1024\), \(d_{\text{ff}} = 4096\), and \(n_{\text{heads}} = 16\), for approximately 311M parameters. Moirai produced probabilistic forecasts by outputting the parameters of a mixture distribution. Patch size was set to auto (selected from \([8, 16, 32, 64, 128]\)), and the inference settings included batch size and the number of samples drawn, with key and value dimension \(d_{kv} = 64\).

We applied Moirai, a TSFM family introduced by Salesforce and trained on large-scale multivariate time series spanning domains such as energy, transport, climate, cloud operations, and finance. Its architecture is based on a masked-encoder transformer with any-variate attention. Multivariate inputs were supplied through \texttt{feat\_dynamic\_real\_dim}, which enabled targets and covariates to be represented jointly and partitioned into patches \citep{woo2024moirai}.

\subsubsection{Chronos}
Chronos-Bolt-Tiny has approximately 9M parameters and is optimised for fast inference. Chronos-Bolt-Mini has approximately 21M parameters and provides a balanced increase in representational capacity. Chronos-Bolt-Small has approximately 48M parameters and offers greater expressive power for modelling non-linear temporal structure. Chronos-Bolt-Base has approximately 205M parameters and enables richer temporal representations at the cost of higher computational demand. Chronos-2 has approximately 120M parameters and serves as a general-purpose TSFM with native support for covariates.

We applied Chronos-Bolt and Chronos-2 as pre-trained TSFMs that produced probabilistic multi-step forecasts. The Chronos models were evaluated using their released capacity presets, and forecasts were generated without updating model weights. Where the implementation supported covariates, we supplied them through the model's supported interface; otherwise, forecasts remained conditioned on target history only. The Chronos models were evaluated under consistent preprocessing and an expanding-window design, and they produced quantile forecasts in a zero-shot setting. Chronos-Bolt can be combined with external covariate regressors, whereas Chronos-2 natively supports all covariate types \citep{ansari2024chronos, ansari2025chronos2}.

\subsubsection{TimesFM}
TimesFM-200M has 200M parameters and 20 layers. TimesFM-500M has 500M parameters and 50 layers. The evaluated configurations used 16 attention heads, patch lengths (input/output) of 32/128, and an embedding dimension of 1280.

We applied TimesFM, a decoder-only pre-trained model designed to operate across variable context and horizon lengths. In the covariate setting, TimesFM generated a base forecast from the target history and then fitted a lightweight linear model to the residuals using the available covariates, thereby yielding a covariate-aware forecast without modifying the core model weights \citep{das2024decoderonlyfoundationmodeltimeseries}.

\subsection{Model evaluation}
In student enrolment forecasting, error measures are the standard means of assessing accuracy \citep{HyndmanKoehler2006}. Mean Absolute Error (MAE) and Mean Squared Error / Root Mean Squared Error (MSE/RMSE), given in Equations~\ref{eq:MAE} and~\ref{eq:mse_rmse}, are among the most commonly reported metrics. MAE was interpretable because it represented the average forecast miss in the original units \citep{WillmottMatsuura2005}. Both metrics compared a forecast \(\hat{y}_t\) with the observed enrolment \(y_t\) across \(T\) periods, but they weighted errors differently: MAE averaged absolute errors, whereas RMSE averaged squared errors and then took the square root, thereby placing greater emphasis on large misses \citep{ChaiDraxler2014}.

\begin{equation}\label{eq:MAE}
    \mathrm{MAE} = \frac{1}{T}\sum_{t=1}^{T} \lvert y_t - \hat{y}_t \rvert
\end{equation}

\begin{equation}\label{eq:mse_rmse}
\begin{aligned}
\mathrm{MSE}  &= \frac{1}{T}\sum_{t=1}^{T}\bigl(y_t - \hat{y}_t\bigr)^2,\\
\mathrm{RMSE} &= \sqrt{\mathrm{MSE}}
             = \sqrt{\frac{1}{T}\sum_{t=1}^{T}\bigl(y_t - \hat{y}_t\bigr)^2}.
\end{aligned}
\end{equation}

To enable cross-series comparison, we also used Symmetric Mean Absolute Percentage Error (SMAPE), given in Equation~\ref{eq:SMAPE}, which rescales the absolute error by the average magnitude of the observed and forecast values. SMAPE is expressed as a percentage and is bounded between \(0\%\) and \(200\%\), although it can become unstable when both \(y_t\) and \(\hat{y}_t\) are close to zero \citep{HyndmanKoehler2006}.

\begin{equation}\label{eq:SMAPE}
    \mathrm{SMAPE} = \frac{100}{T}\sum_{t=1}^{T} \frac{2\lvert y_t - \hat{y}_t\rvert}{\lvert y_t\rvert + \lvert \hat{y}_t\rvert}
\end{equation}

\subsection{Evaluating probabilistic forecasts and calibration diagnostics}
We evaluated probabilistic forecasts using proper, and ideally strictly proper, scoring rules. These loss functions align incentives by ensuring that the expected score is minimised only when the forecaster reports its true predictive distribution \citep{waghmare2025properscoringrulesestimation}. Building on the classical foundation for evaluating binary events, we relied on the broader theory of strictly proper scoring rules applicable to both discrete and continuous outcomes \citep{MathesonWinkler1976}. This mathematical rigour was important because it ensured that improvements in the metrics reflected genuine gains in predictive performance \citep{GneitingBalabdaouiRaftery2007}.

For continuous targets, we used Continuous Ranked Probability Score (CRPS), which is reported alongside, and contrasted with, the logarithmic score \citep{ArnoldWalzZiegelGneiting2024}. CRPS evaluates global fit by comparing the entire predictive Cumulative Distribution Function (CDF), denoted by \(F\), with the realised value \(y\), as shown in Equation~\ref{eq:CRPS}:
\begin{equation}
\label{eq:CRPS}
\mathrm{CRPS}(F, y) = \int_{-\infty}^{\infty} \left( F(z) - \mathbb{1}\{y \leq z\} \right)^2 \, dz
\end{equation}

Beyond aggregate scores, we assessed distributional reliability using Probability Integral Transform (PIT). Under probabilistic calibration, realisations mapped through the predicted CDF should behave as uniform noise and follow a uniform distribution \(\mathcal{U}[0,1]\) \citep{Rosenblatt1952}. PIT diagnostics were used to visually detect bias, overconfidence, and under-dispersion, following established practice in econometrics and meteorology \citep{DieboldGuntherTay1998}. In addition to significance testing, we reported effect sizes to quantify the \emph{magnitude} of improvement in a way that remained interpretable under data sparsity. For point-forecast accuracy, we computed paired differences relative to a reference baseline.

\clearpage

\subsection{Experiment setup}
We implemented a reproducible forecasting pipeline with strict vintage control (Figure~\ref{fig:experiment_pipeline}). Processing operated at the yearly level; we aligned all series to a common year index and standardised features within each training window to prevent information leakage. Model evaluation followed an expanding-window backtest for each annual forecast origin.

\subsubsection{Task definition}
We studied annual year-ahead forecasting of new enrolments in a higher-education institutional planning setting. New enrolments referred to students commencing their studies at the institution in the given academic year. The target series was observed at an annual cadence and was disaggregated into institutional cohorts. The annual sampling and short history imposed severe data sparsity, motivating the use of zero-shot TSFMs and an external covariate.

\subsubsection{Backtesting protocol}
We adopted an expanding-window backtest. At each forecast origin year $t$, models produced a one-year-ahead forecast for $t{+}1$ using only information that would have been available at time $t$. All external covariates were lagged and vintage-aligned to the forecast origin to avoid leakage. This protocol was designed to reflect a decision-time institutional workflow: forecasts were generated annually for planning, and covariate availability was constrained by publication timing and platform release schedules.

\subsubsection{Models and baselines}
We benchmarked multiple TSFMs in a strict zero-shot configuration. We compared against a reference baseline suitable for annual planning. For TSFMs that output predictive distributions, we evaluated both point and probabilistic quality. Model variants were treated as a controlled factor to assess whether covariate conditioning interacted with model capacity under sparsity.

\subsubsection{Covariate sets and ablation design}
External information was introduced via a compact, leakage-safe covariate set with explicit availability rules. We structured the experiment as an ablation grid over covariate regimes. This design clarified, under annual sparsity, when leakage-safe covariates improved zero-shot TSFMs and when they introduced brittle conditioning behaviours.

\subsubsection{Evaluation metrics and diagnostics}
We reported both point and probabilistic forecast performance. For point accuracy, we used error metrics, supplemented with scale-normalised error measures where appropriate to support comparisons across cohorts and periods. To summarise practical magnitude, we reported paired differences and effect sizes relative to a reference baseline, quantifying the size of improvements in a way that remained interpretable under short, annual histories.

\subsubsection{Reporting and reproducibility}
To support auditability, we reported covariate availability rules, lag choices, and feature-engineering steps alongside the backtesting configuration. Where data was restricted, we specified an explicit data-availability statement and provided sufficient procedural detail to allow reproduction of the evaluation and covariate construction steps on other institutions’ data.

\begin{figure*}[t!]
\centering
\captionsetup{font=small}
\begin{tikzpicture}[
  font=\small,
  node distance=6mm and 10mm,
  >=Latex,
  box/.style={rounded corners, draw, align=left, inner sep=5pt, text width=0.86\linewidth},
  mini/.style={rounded corners, draw, align=left, inner sep=4pt, text width=0.34\linewidth},
  arrow/.style={->, line width=0.6pt}
]

\node[box] (in) {\textbf{Inputs (annual)}: new enrolments (by cohort) + leakage-safe covariates (Google Trends, IOCI).\\
\textbf{Vintage rule}: at origin $t$, use only information available at $t$; covariates vintage-aligned.};

\node[box, below=of in] (prep) {\textbf{Preprocess}: align to common year index; feature engineering; \textbf{standardise within each training window} (no leakage).};

\node[box, below=of prep] (bt) {\textbf{Expanding-window backtest}: for each origin $t$, train on $\le t$ and forecast $t{+}1$.};

\node[box, below=of bt] (exp) {\textbf{Experiment factors}: zero-shot TSFMs + annual-planning baseline, controlled variants to test covariate conditioning under sparsity.
\\ Covariates:\\
(A) No Covariate\\
(B) Google Trends\\
(C) IOCI\\
};



\node[box, below=of exp] (out) {\textbf{Outputs}: point errors + probabilistic quality + paired differences (effect sizes).\\
\textbf{Reproducibility}: lags, availability rules, and data-availability statement.};

\draw[arrow] (in) -- (prep);
\draw[arrow] (prep) -- (bt);
\draw[arrow] (bt) -- (exp);
\draw[arrow] (exp) -- (out);

\end{tikzpicture}

\caption{Experiment workflow.}
\label{fig:experiment_pipeline}
\end{figure*}
\clearpage

\section{Experimental results and analysis}
This section reports forecasting accuracy across model classes under different input settings. We summarise performance using standard error metrics.

\subsection{Forecast analysis of model evaluation results}
Table~\ref{tab:all_results} summarises per-cohort and per-condition rankings by ordering models within each covariate setting according to four error measures (MAE, RMSE, SMAPE, and MAPE), and then averaging these ranks across metrics and forecast horizons. This yields an interpretable summary score for each model within each cohort that reflects overall forecasting performance.

\begin{sidewaystable*}[p]
\centering
\captionsetup{font=footnotesize,skip=2pt}
\caption{Model comparisons metrics all results for domestic and international students enrolment (good = green, orange = moderate, red = bad).}
\label{tab:all_results}

\scriptsize
\setlength{\tabcolsep}{2.2pt}%
\renewcommand{\arraystretch}{0.78}%

\begin{adjustbox}{max width=\textheight, max totalheight=0.93\textwidth, keepaspectratio}
\begin{tabular}{llccccc}
\toprule
\textbf{Covariates} & \textbf{Model} & \textbf{MAE} & \textbf{RMSE} & \textbf{SMAPE} & \textbf{MAPE} & \textbf{MAE Rank} \\
\midrule

\multicolumn{7}{c}{\textbf{Domestic student}}\\
\midrule

\multirow{12}{*}{No Covariate}
& Chronos-Bolt-Tiny   & \good{358.1} & \good{455.3} & \good{6.1} & \good{6.3} & \good{1}  \\
& Persistence           & \good{371.5} & \good{456.6} & \med{6.4}  & \med{6.7}  & \good{2}  \\
& Chronos-Bolt-Small  & \good{372.6} & \med{480.3}  & \good{6.3} & \good{6.5} & \good{3}  \\
& Chronos-Bolt-Mini   & \med{376.5}  & \good{461.5} & \med{6.4}  & \med{6.7}  & \med{4}   \\
& Chronos-2             & \med{386.3}  & \med{481.9}  & \med{6.6}  & \med{6.9}  & \med{5}   \\
& TimesFM-500          & \med{395.7}  & \med{479.9}  & \med{6.8}  & \med{7.1}  & \med{6}   \\
& TimesFM-200          & \med{397.3}  & \med{511.0}  & \med{6.8}  & \med{7.2}  & \med{7}   \\
& Chronos-Bolt-Base   & \med{406.7}  & \med{491.8}  & \med{7.0}  & \med{7.3}  & \med{8}   \\
& Moirai-Small         & \bad{506.2}  & \bad{621.6}  & \bad{8.6}  & \bad{9.1}  & \bad{9}   \\
& ARIMA                 & \bad{508.8}  & \bad{614.9}  & \bad{8.4}  & \bad{8.9}  & \bad{10}  \\
& Moirai-Large         & \bad{519.5}  & \bad{626.4}  & \bad{8.8}  & \bad{9.3}  & \bad{11}  \\
& Moirai-Base          & \bad{531.5}  & \bad{656.8}  & \bad{9.1}  & \bad{9.6}  & \bad{12}  \\

\midrule

\multirow{12}{*}{IOCI}
& Chronos-Bolt-Tiny   & \good{321.3} & \good{431.9} & \good{5.5} & \good{5.8} & \good{1}  \\
& Chronos-Bolt-Small  & \good{324.0} & \good{423.6} & \good{5.5} & \good{5.8} & \good{2}  \\
& Chronos-Bolt-Base   & \good{347.0} & \good{438.3} & \good{6.0} & \good{6.3} & \good{3}  \\
& Chronos-Bolt-Mini   & \med{348.5}  & \med{431.2}  & \med{6.0}  & \med{6.3}  & \med{4}   \\
& Persistence           & \med{371.5}  & \med{456.6}  & \med{6.4}  & \med{6.7}  & \med{5}   \\
& Chronos-2             & \med{387.7}  & \med{504.7}  & \med{6.6}  & \med{7.0}  & \med{6}   \\
& Moirai-Small         & \med{496.6}  & \med{611.0}  & \med{8.4}  & \med{9.0}  & \med{7}   \\
& TimesFM-500          & \med{509.8}  & \med{583.3}  & \med{9.0}  & \med{9.4}  & \med{8}   \\
& Moirai-Base          & \bad{528.5}  & \bad{662.9}  & \bad{9.0}  & \bad{9.7}  & \bad{9}   \\
& TimesFM-200          & \bad{532.4}  & \bad{618.2}  & \bad{9.4}  & \bad{9.9}  & \bad{10}  \\
& Moirai-Large         & \bad{628.9}  & \bad{843.3}  & \bad{10.3} & \bad{11.3} & \bad{11}  \\
& ARIMAX                 & \bad{698.1}  & \bad{786.1}  & \bad{12.1} & \bad{13.0} & \bad{12}  \\

\midrule
\multicolumn{7}{c}{\textbf{International student}}\\
\midrule

\multirow{12}{*}{No Covariate}
& Moirai-Small         & \good{168.8} & \good{180.9} & \good{14.7} & \good{15.0} & \good{1}  \\
& Moirai-Base          & \good{183.6} & \good{197.5} & \good{16.9} & \good{17.0} & \good{2}  \\
& Chronos-Bolt-Small  & \good{189.3} & \good{200.2} & \good{17.8} & \good{18.0} & \good{3}  \\
& Chronos-Bolt-Mini   & \med{200.5}  & \med{218.8}  & \med{19.0}  & \med{18.7}  & \med{4}   \\
& Chronos-2             & \med{201.0}  & \med{210.0}  & \med{18.3}  & \med{18.8}  & \med{5}   \\
& Chronos-Bolt-Tiny   & \med{205.2}  & \med{216.4}  & \med{19.1}  & \med{19.6}  & \med{6}   \\
& Chronos-Bolt-Base   & \med{208.9}  & \med{222.7}  & \med{19.4}  & \med{19.1}  & \med{7}   \\
& Persistence           & \med{209.6}  & \med{225.5}  & \med{19.2}  & \med{20.0}  & \med{8}   \\
& ARIMA                 & \bad{217.8}  & \bad{234.3}  & \bad{19.2}  & \bad{19.9}  & \bad{9}   \\
& Moirai-Large         & \bad{222.4}  & \bad{231.0}  & \bad{20.0}  & \bad{20.5}  & \bad{10}  \\
& TimesFM-500          & \bad{226.6}  & \bad{248.3}  & \bad{20.8}  & \bad{20.7}  & \bad{11}  \\
& TimesFM-200          & \bad{236.4}  & \bad{258.0}  & \bad{21.2}  & \bad{21.3}  & \bad{12}  \\

\midrule

\multirow{12}{*}{Google Trends}
& Moirai-Base          & \good{181.3} & \good{190.0} & \good{16.6} & \good{17.6} & \good{1}  \\
& Moirai-Small         & \good{181.7} & \good{192.6} & \good{15.7} & \good{16.4} & \good{2}  \\
& Chronos-Bolt-Base   & \good{190.8} & \good{215.6} & \good{17.8} & \good{17.4} & \good{3}  \\
& Chronos-Bolt-Mini   & \med{196.4}  & \med{215.4}  & \med{18.0}  & \med{18.2}  & \med{4}   \\
& Chronos-Bolt-Small  & \med{202.6}  & \med{225.1}  & \med{18.9}  & \med{19.1}  & \med{5}   \\
& Persistence           & \med{209.6}  & \med{225.5}  & \med{19.2}  & \med{20.0}  & \med{6}   \\
& Chronos-2             & \med{212.1}  & \med{228.7}  & \med{19.3}  & \med{19.7}  & \med{7}   \\
& Chronos-Bolt-Tiny   & \med{212.3}  & \med{235.4}  & \med{20.7}  & \med{20.3}  & \med{8}   \\
& Moirai-Large         & \bad{230.3}  & \bad{241.8}  & \bad{20.9}  & \bad{21.3}  & \bad{9}   \\
& TimesFM-200          & \bad{232.6}  & \bad{269.4}  & \bad{21.0}  & \bad{21.3}  & \bad{10}  \\
& TimesFM-500          & \bad{238.6}  & \bad{273.0}  & \bad{21.6}  & \bad{22.2}  & \bad{11}  \\
& ARIMAX                 & \bad{364.1}  & \bad{424.3}  & \bad{33.8}  & \bad{38.2}  & \bad{12}  \\

\bottomrule
\end{tabular}
\end{adjustbox}
\end{sidewaystable*}

\subsubsection{Domestic cohort}
\begin{figure*}[t!]
\centering
\captionsetup{font=small}
\captionsetup[subfigure]{font=small, labelfont=bf}

\newcommand{\covwD}{0.8\linewidth}
\newcommand{\fcwD}{0.8\linewidth}

\begin{subfigure}[t]{\textwidth}
  \centering
  \includegraphics[width=0.8\textwidth]{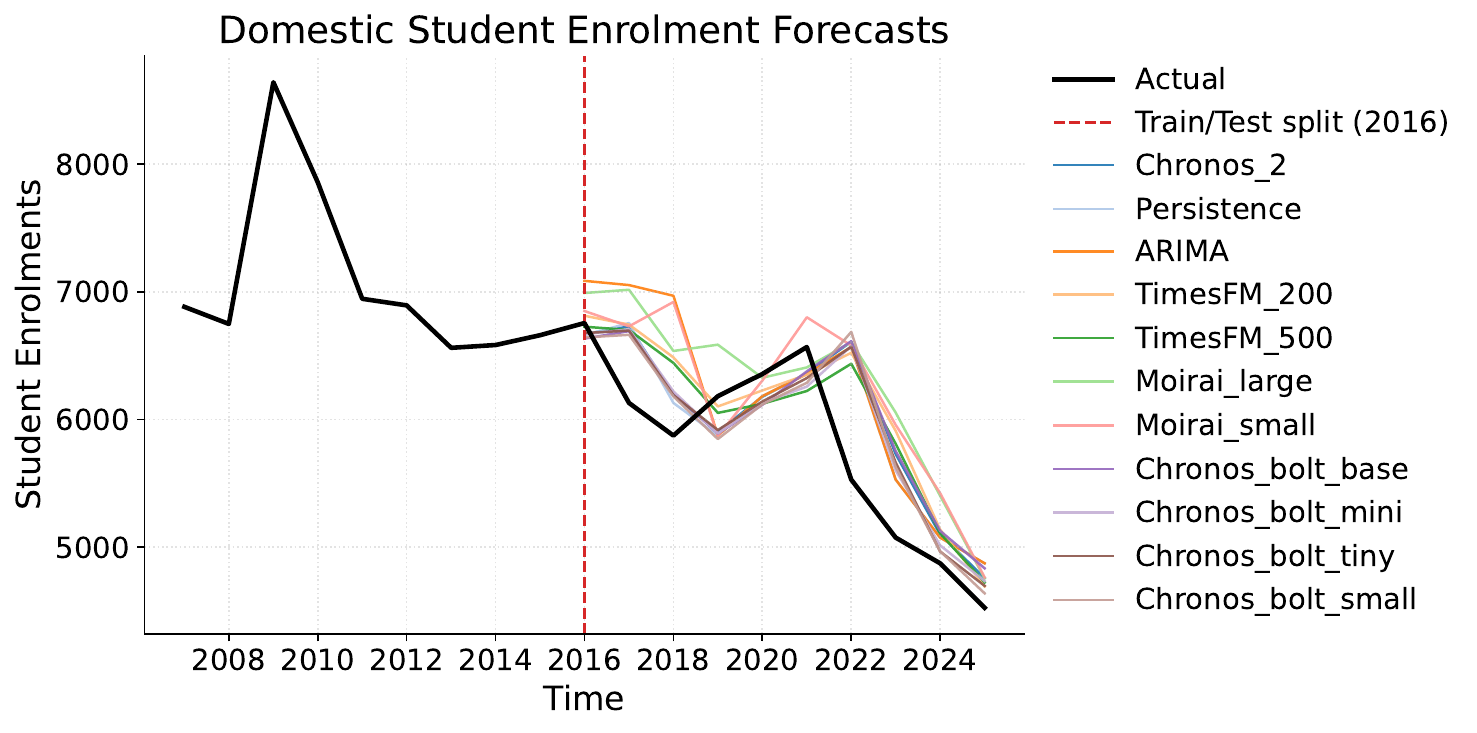}
  \caption{No covariate.}
  \label{fig:domestic-no-cov}
\end{subfigure}

\vspace{2mm}

\begin{subfigure}[t]{\textwidth}
  \centering
  \begin{minipage}[t]{\covwD}
    \vspace{0pt}\centering
    \includegraphics[width=\linewidth]{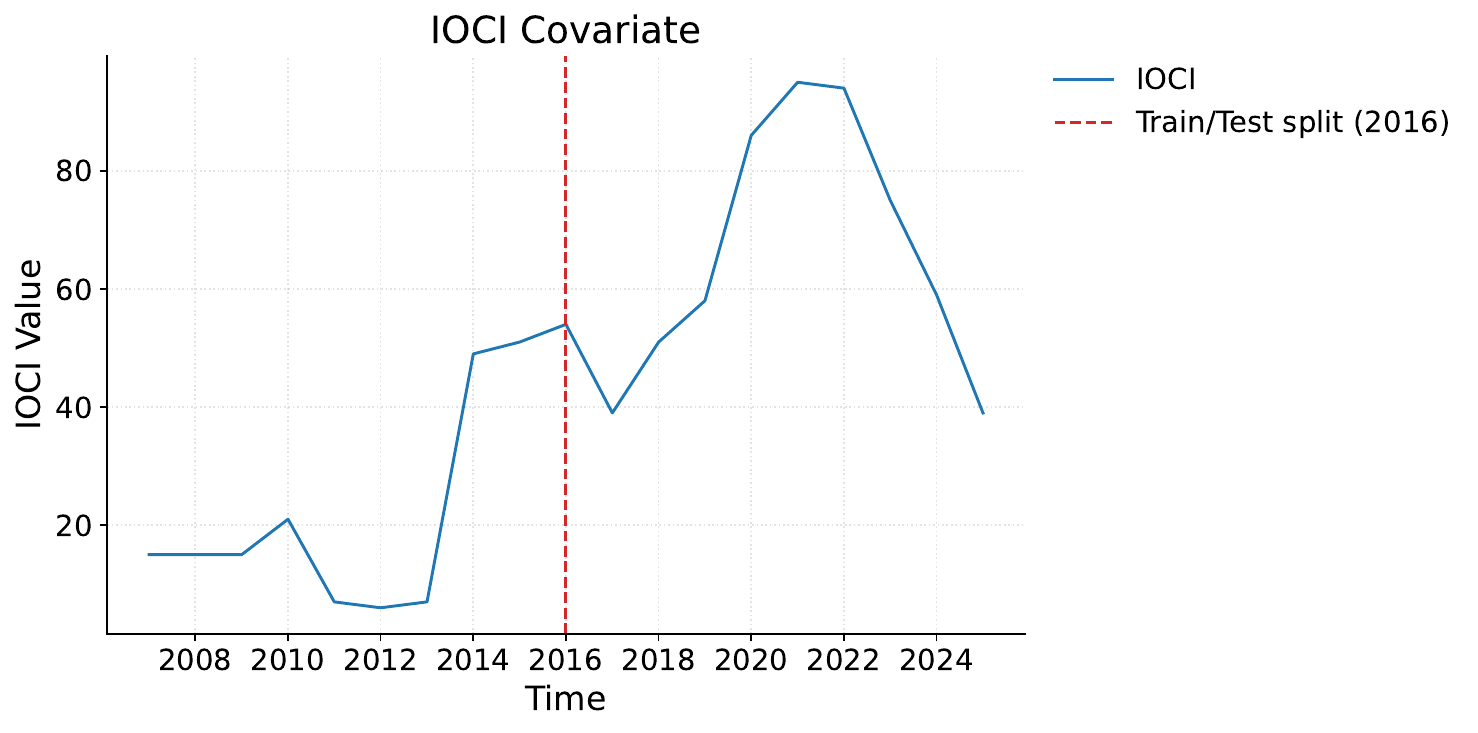}
  \end{minipage}\hfill
  \begin{minipage}[t]{\fcwD}
    \vspace{0pt}\centering
    \includegraphics[width=\linewidth]{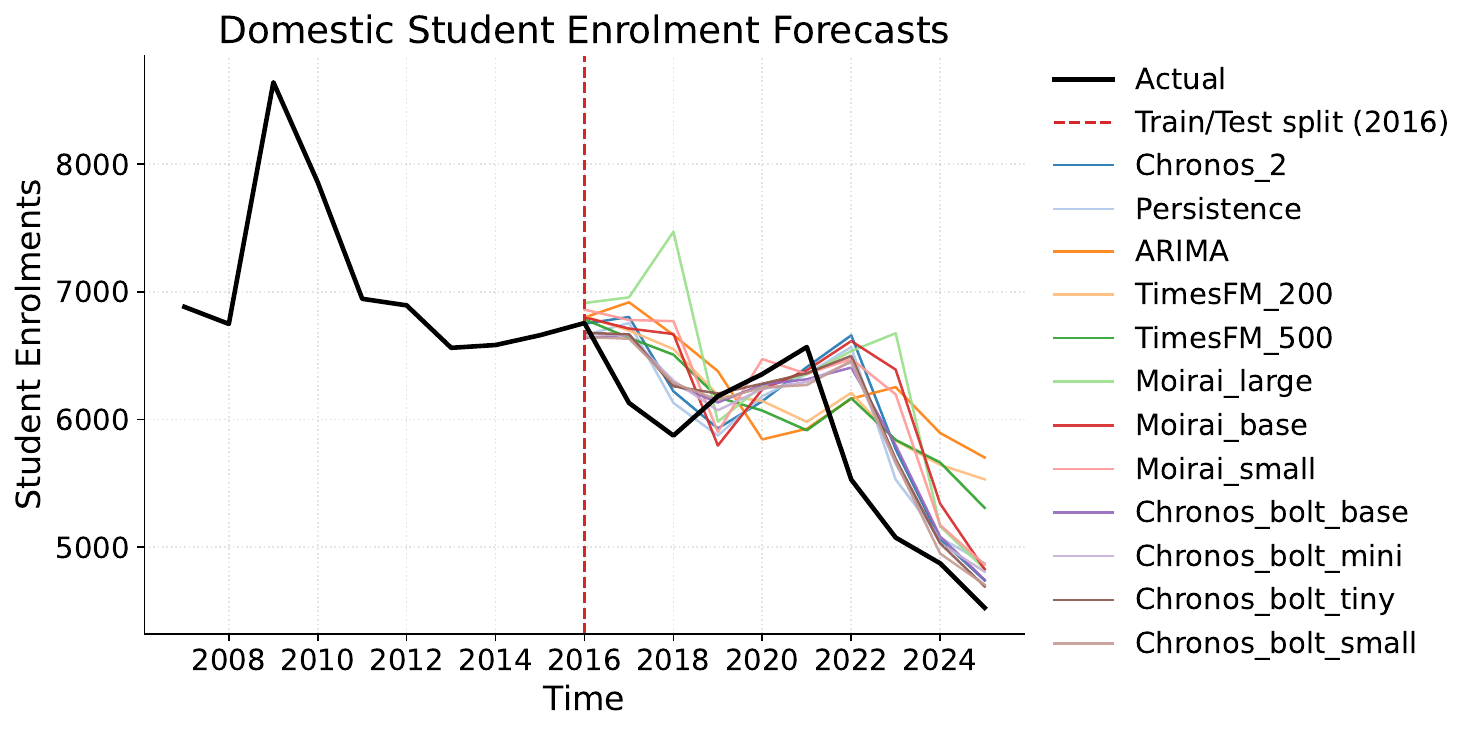}
  \end{minipage}
  \caption{With the IOCI covariate (top: covariate; bottom: forecasts).}
  \label{fig:domestic-ioci}
\end{subfigure}

\caption{Actual and forecast annual new domestic-student enrolments. Panel (a) shows unconditional forecasts. Panel (b) shows the IOCI covariate and the covariate-conditioned forecasts. The red dashed line marks the forecast origin year.}
\label{fig:domestic-results}
\end{figure*}

Figure~\ref{fig:domestic-results} shows the results for new domestic enrolments. Adding calibrated IOCI substantially improved the performance of the Chronos-Bolt family, with consistent gains across the error metrics. This change reordered the relative performance ranking in favour of the Chronos variants. By contrast, Persistence dropped in rank, while TimesFM and ARIMAX deteriorated substantially when the IOCI covariate was introduced. These results suggest that calibrated IOCI improved performance primarily for model families with stable covariate-conditioning mechanisms.

\subsubsection{International cohort}
\begin{figure*}[ht!]
\centering
\newcommand{\covw}{0.8\linewidth}
\newcommand{\fcw}{0.8\linewidth}
\begin{subfigure}{\textwidth}
  \centering
  \includegraphics[width=0.8\textwidth]{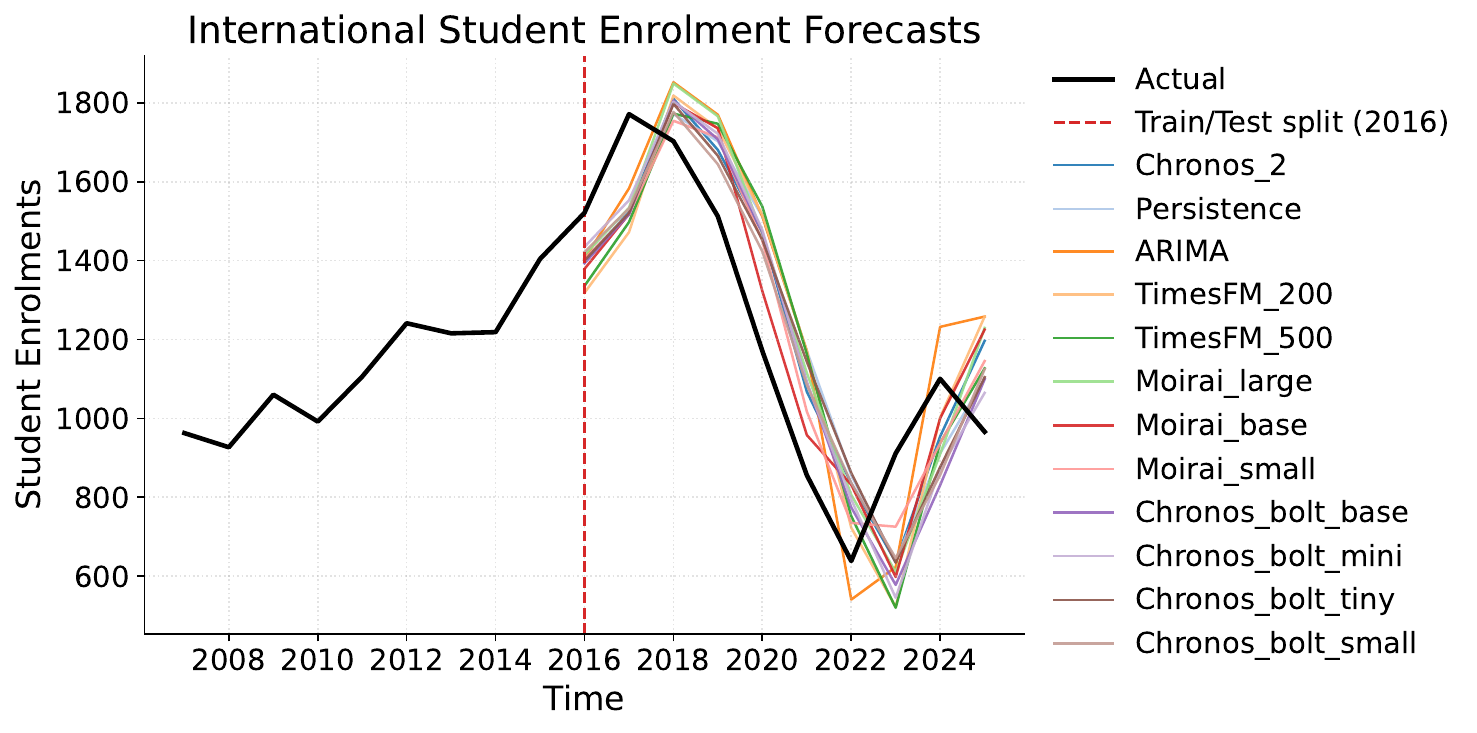}
  \caption{No covariate.}
\end{subfigure}

\vspace{2mm}

\begin{subfigure}{\textwidth}
\centering
  \begin{minipage}[t]{\covw}
    \vspace{0pt}\centering
    \includegraphics[width=\linewidth]{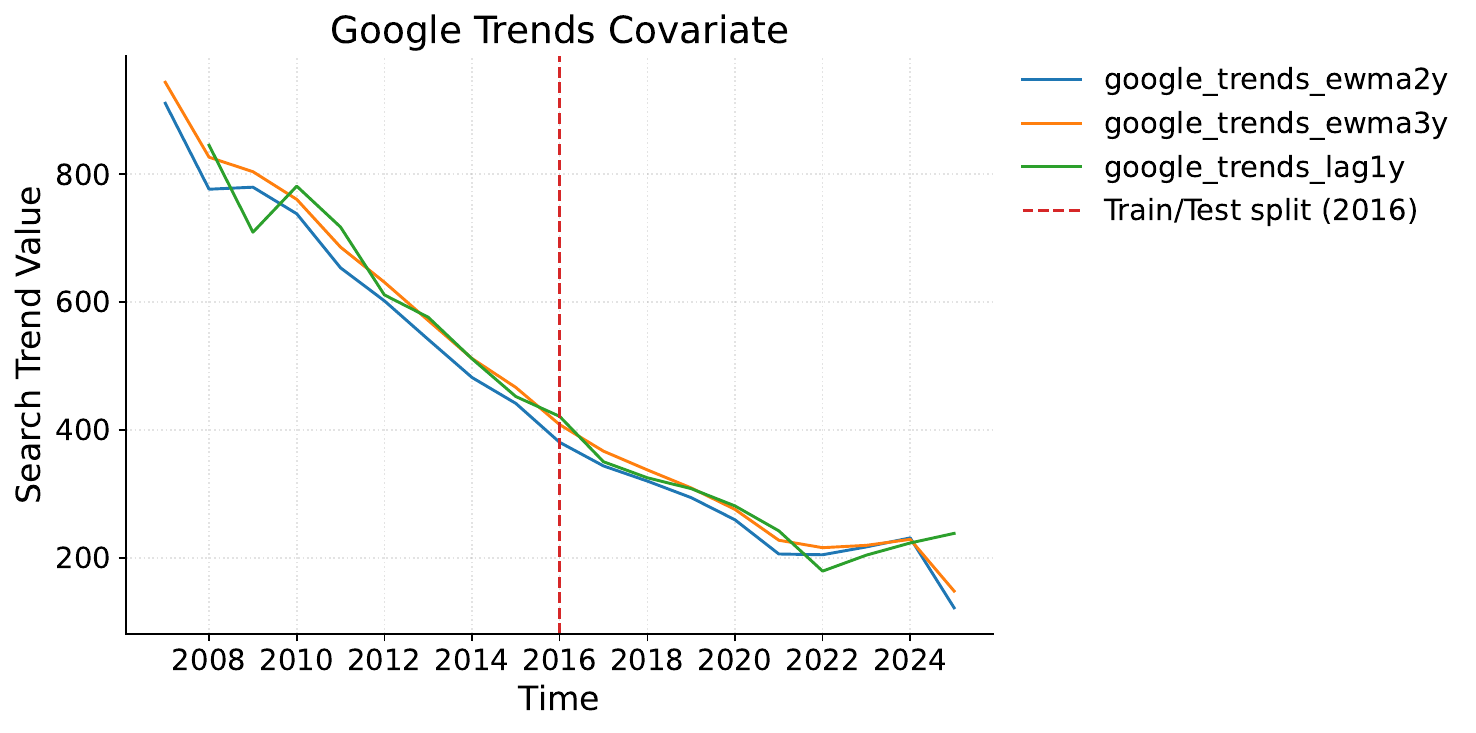}
  \end{minipage}\hfill
  \begin{minipage}[t]{\fcw}
    \vspace{0pt}\centering
    \includegraphics[width=\linewidth]{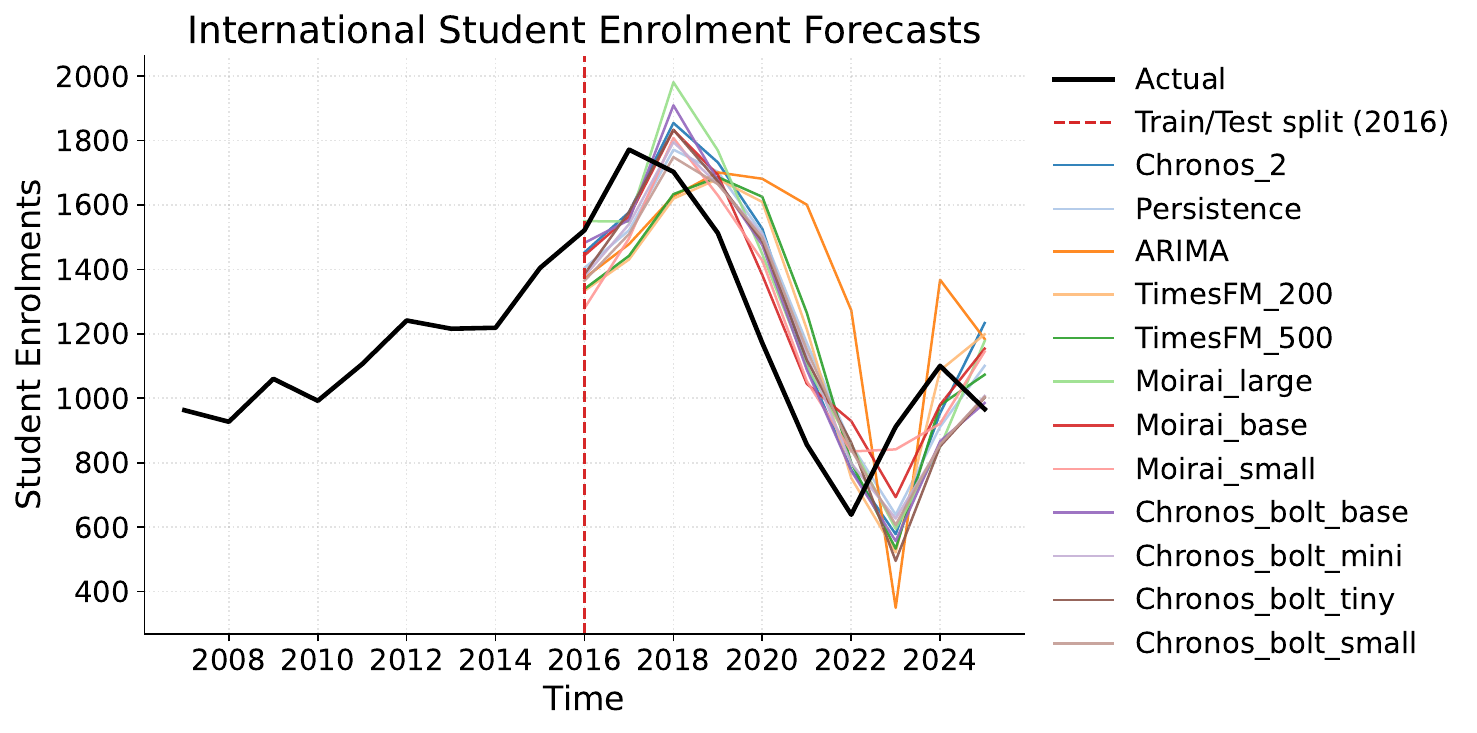}
  \end{minipage}
  \caption{With feature-engineered Google Trends as covariates (top: covariates; bottom: forecasts).}
\end{subfigure}
\caption{Actual and forecast annual new international-student enrolments. Panel (a) shows unconditional forecasts. Panel (b) shows the feature-engineered Google Trends and the covariate-conditioned forecasts. The red dashed line marks the forecast origin year.}
\label{fig:international-results}
\end{figure*}

Figure~\ref{fig:international-results} shows the results for new international enrolments. In the no-covariate setting, Moirai-Small achieved the lowest errors and ranked first overall. Introducing feature-engineered Google Trends did not produce uniform gains in the Trends-augmented setting. Moirai-Base became the top-ranked model, followed closely by Moirai-Small, while the strongest Chronos variant under the Trends setting was Chronos-Bolt-Base. This indicates that feature-engineered Google Trends acted as a selective demand-proxy signal that was helpful for some model families, improving Moirai-Base and Chronos-Bolt-Base, but degrading others, most notably Moirai-Small, relative to its no-covariate performance.

\subsubsection{Impact of covariates}
The results indicate a clear interaction between covariates and model class, with engineered features derived from Google Trends and IOCI improving accuracy for several models. For domestic enrolments, IOCI acted as a stabilising covariate for the Chronos-Bolt family, delivering consistent improvements and preserving Chronos-Bolt-Tiny as the best-performing model. However, ARIMAX and TimesFM did not benefit from this covariate and deteriorated when IOCI was introduced, suggesting a limited capacity to exploit the signal without amplifying variance.

For international enrolments, the impact of Google Trends was mixed. Under Trends conditioning, Moirai-Base ranked first, and Chronos-Bolt-Base and Chronos-Bolt-Mini improved relative to their no-covariate counterparts. Operationally, this suggests that Google Trends features should be treated as an optional covariate. They can improve performance for some model families, but they are not uniformly beneficial.

\clearpage

\subsubsection{Effect size and probabilistic calibration}
{\small 
\setlength{\tabcolsep}{3.5pt}%
\renewcommand{\arraystretch}{0.90}%
\setlength{\LTpre}{0pt}%
\setlength{\LTpost}{0pt}%
\setlength{\aboverulesep}{0pt}%
\setlength{\belowrulesep}{0pt}
\begin{table}[ht]
\centering
\caption{Effect Size and CRPS Summary}
\label{tab:effect_size_CRPS}
\begin{tabular}{lrrrrrr}
\toprule
Cohort & Covariates & Model & $\Delta${MAE} &  CRPS (80\%) & CRPS (95\%)  \\
\midrule
Domestic & None & Chronos-Bolt-Tiny & 13.5 & 254.7 & 284.4 \\
\midrule
Domestic & IOCI & Chronos-Bolt-Tiny  & 50.22 & 239.7 & 269.5 \\
Domestic & IOCI & Chronos-Bolt-Small & 47.49 & 243.9 & 275.1 \\
Domestic & IOCI & Chronos-Bolt-Mini  & 23.04 & 249.5 & 281.2 \\
Domestic & IOCI & Chronos-Bolt-Base  & 24.47 & 256.7 & 293.3 \\
\midrule
International & None & Moirai-Small        & 40.8  & 105.4 & 113.1 \\
International & None & Moirai-Base         & 26.0  & 117.7 & 129.3 \\
International & None & Chronos-Bolt-Small  & 20.3  & 120.1 & 129.8 \\
International & None & Chronos-Bolt-Mini   & 9.1   & 127.9 & 138.8 \\
International & None & Chronos-2           & 8.6   & 127.7 & 138.9 \\
International & None & Chronos-Bolt-Tiny   & 4.4   & 130.6 & 143.2 \\
International & None & Chronos-Bolt-Base   & 0.7   & 135.9 & 149.6 \\
\midrule
International & Google Trends & Moirai-Base & 28.3 & 119.1 & 129.6 \\
International & Google Trends & Moirai-Small & 27.9 & 119.9 & 130.5 \\
International & Google Trends & Chronos-Bolt-Base & 18.8 & 135.7 & 154.1 \\
International & Google Trends & Chronos-Bolt-Mini & 13.3 & 131.1 & 145.8 \\
International & Google Trends & Chronos-Bolt-Small  & 7.0  & 144.6 & 166.8 \\
\bottomrule
\end{tabular}
\end{table}
}

Table \ref{tab:effect_size_CRPS} provides a compact summary of both practical accuracy gains and probabilistic forecast quality across cohorts, covariate settings, and TSFMs. The table is organised by cohort, then by the type of external information available to the model, and finally by forecasting model configuration. For each row, $\Delta\mathrm{MAE} = \mathrm{MAE}{\text{Persistence}} - \mathrm{MAE}{\text{Model}}$, which summarises the reduction in point error relative to the reference baseline and serves as an effect-size indicator of improvement. In addition, CRPS (80\%) and CRPS (95\%) summarise probabilistic forecast quality, with lower values indicating better uncertainty-aware predictions. In our setting, several forecasters provide discrete predictive quantiles, so we use a quantile-based approximation to CRPS based on central prediction intervals. CRPS (80\%) is computed from the quantiles defining the central 80\% interval, while CRPS (95\%) is computed from the quantiles defining the central 95\% interval. These probabilistic measures complement point-error metrics under annual data sparsity. The results indicate that covariates can improve both point and probabilistic performance, although the consistency of those gains depends on the cohort, model family, and covariate-conditioning design under small-sample backtesting.

\begin{figure}[ht!]
\begin{subfigure}{\textwidth}
  \centering
  \includegraphics[width=0.9\textwidth]{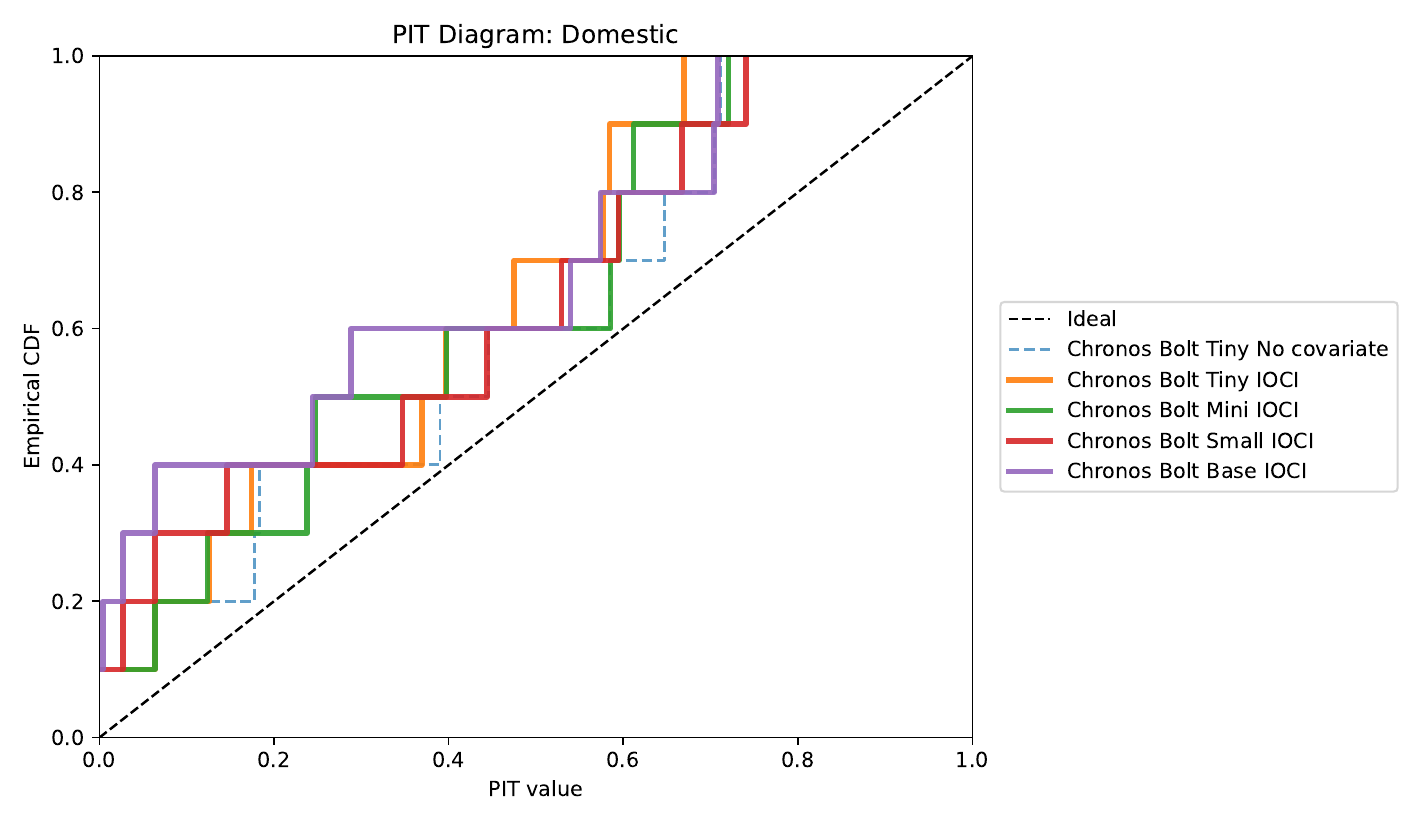}
\end{subfigure}
\begin{subfigure}{\textwidth}
  \centering
  \includegraphics[width=0.9\textwidth]{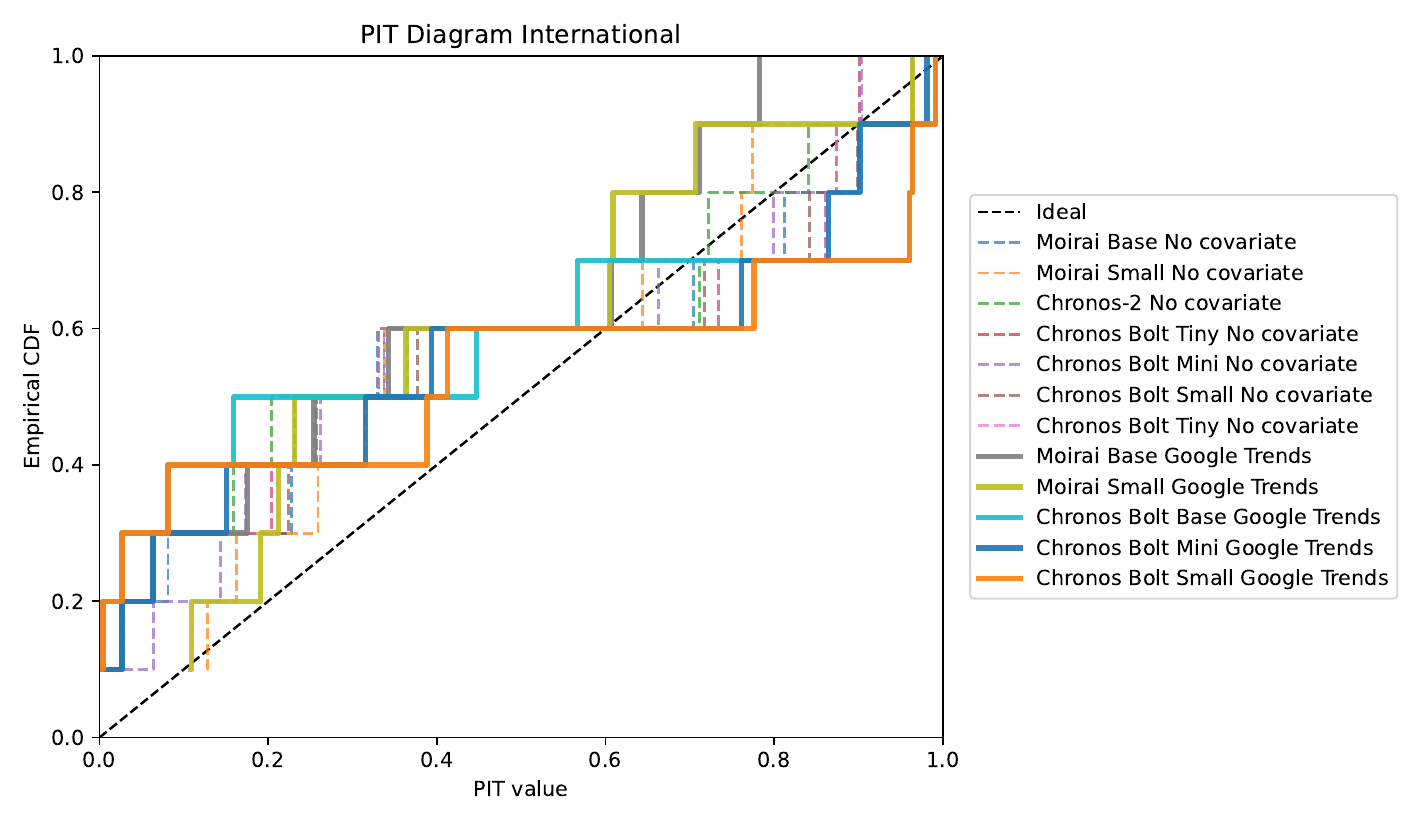}
\end{subfigure}
\caption{Empirical CDF of PIT values.}
  \label{fig:PIT}
\end{figure}

Figure~\ref{fig:PIT} reports a standard diagnostic for distributional calibration. For each forecast origin $t$, the PIT value is defined as $u_t = F_t(y_t)$, that is, the predictive CDF evaluated at the realised outcome. Under ideal calibration, $u_t \sim \mathrm{Uniform}(0,1)$, and the empirical CDF of $\{u_t\}$ follows the 45$^\circ$ reference line.

In the domestic cohort, the no-covariate setting is compared with IOCI-conditioned Chronos-Bolt variants. The IOCI-conditioned curves lie closer to the diagonal across much of the PIT range, indicating improved probabilistic calibration. While departures from the reference line remain, the overall shift towards the diagonal is consistent with IOCI providing additional regime information that improves the alignment between predictive distributions and realised outcomes.

In the international cohort, Google Trends configurations generally track the reference line more closely than their no-covariate counterparts, through the middle-to-upper PIT range. This pattern is consistent with the features improving the allocation of probability mass during periods of strong demand variation. However, several model-covariate combinations still exhibit pronounced deviations, underscoring that calibration gains are configuration-dependent.

\subsection{Forecast analysis of new domestic student enrolments}
The domestic tertiary sector exhibits gradual year-to-year aggregate movements. This inherent stability renders simple extrapolative benchmarks competitive, given the modest domestic growth observed in the sector in recent years. In this analysis, we evaluated forecasts of new domestic enrolments using an expanding-window backtesting design, comparing classical baselines against zero-shot TSFMs under varying covariate specifications.

The time series for new domestic enrolments is characterised by relatively smooth medium-run evolution. Consistent with this pattern, the naive Persistence baseline serves as a rigorous benchmark. Among the TSFMs, the compact Chronos-Bolt variants demonstrate superior performance, suggesting that they regularise effectively within limited historical windows. In the absence of external covariates, Chronos-Bolt-Tiny is followed closely by the Persistence baseline and Chronos-Bolt-Small. The remaining Chronos and TimesFM variants occupy an intermediate position, whereas the Moirai variants and ARIMA exhibit higher errors.

The incorporation of IOCI reveals a heterogeneous response. The Chronos-Bolt family derives substantial benefit from this covariate: Chronos-Bolt-Tiny and Chronos-Bolt-Small improved under IOCI, with $\Delta$MAE values of 50.22 and 47.49 relative to Persistence, respectively. Conversely, TimesFM and ARIMAX degraded sharply after the introduction of the index. This divergence suggests that while IOCI contains a stabilising, regime-aware signal, its utility is contingent on the model's conditioning mechanism. The additional covariate channel can amplify misspecification and induce brittle extrapolation. In the expanding-window backtests, IOCI improved domestic forecasts for specific TSFM configurations but did not improve international forecasts consistently; we therefore retained IOCI only in settings where it provided an incremental signal.

Across the evaluated configurations, Chronos-Bolt-Tiny provided the most favourable accuracy-stability trade-off in the no-covariate setting. When a leakage-safe institutional context signal was available, conditioning Chronos-Bolt-Tiny and, in some settings, Chronos-Bolt-Small on IOCI conferred additional reductions in error. Overall, Chronos-Bolt-Tiny showed stable performance. When a leakage-safe institutional context signal is available, Chronos-Bolt-Tiny and Chronos-Bolt-Small conditioned on IOCI are recommended for improved accuracy and more regime-aware tracking. ARIMAX requires careful diagnostics and can degrade materially under short samples and structural change. The reductions in point errors for the Chronos variants are consistent with what would be expected under annual sampling and short evaluation windows relative to the Persistence baseline.


\subsection{Forecast analysis of new international student enrolments}
New international student enrolments exhibit heightened sensitivity to exogenous shocks, with abrupt deviations driven by border settings, visa policy, geopolitical risk, and shifts in perceived destination attractiveness. Consistent with this macro-sensitivity, international commencing enrolments rebounded in 2024 relative to 2023, although they remained below pre-pandemic levels, underscoring ongoing structural uncertainty.

The international time series displays greater variance and more pronounced turning-point behaviour. Consequently, purely autoregressive baselines risk lagging during rapid upswings or drawdowns. In such settings, model efficacy is enhanced by architectures with strong cross-domain priors, such as TSFMs, and by the inclusion of leakage-safe external proxies that capture demand intent. However, the utility of exogenous signals depends on the downstream model family and the stability of the covariate transformations.

In the univariate setting, excluding covariates, Moirai-Small demonstrated superior accuracy, followed by Moirai-Base and Chronos-Bolt-Small. While classical baselines remained competitive, the TimesFM variants and ARIMA trailed the top-performing TSFMs. This performance hierarchy suggests that for international demand, masked-encoder TSFMs (Moirai) provide a strong default prior even in the absence of explicit exogenous inputs, plausibly reflecting their ability to generalise multiscale structure learned during pre-training.

Feature-engineered Google Trends induced a material shift in the performance hierarchy. Moirai-Base ascended to the top rank, whereas Moirai-Small ceded its lead. This divergence in response is informative. While some Chronos variants were able to leverage the Trends signal, ARIMAX performance deteriorated, indicating sensitivity to covariate misspecification and instability under short samples.

These results suggest that the benefit of which is contingent on the structural compatibility between covariate conditioning and the forecasting model. Despite these gains for the best-performing models, the improvements should be interpreted cautiously given the annual sampling frequency and the limited number of forecast origins.


\section{Discussion}
\subsection{Forecasting design for annual institutional planning under severe data sparsity}
For \textbf{RQ1}, the findings demonstrate that state-of-the-art TSFMs provide zero-shot enrolment forecasting in institutional settings constrained by short historical records and privacy restrictions. Across both domestic and international cohorts, zero-shot TSFMs match or outperform classical benchmarks without the operational overhead of task-specific fine-tuning. 

The results also show that the preferred forecasting design depends on the structure of the planning series. For the domestic cohort, where annual movements were relatively smooth, compact Chronos-Bolt variants offered the most favourable balance between stability and responsiveness. In this setting, a smaller model capacity appeared sufficient to exploit short-run structure without amplifying noise. For the international cohort, which exhibited greater volatility and stronger turning-point behaviour, Moirai variants performed strongly in the no-covariate setting, suggesting that masked-encoder architectures offer a useful prior for more shock-sensitive demand series.

From an operational research perspective, the implication is that model choice should be treated as part of a broader planning-system design. In data-sparse annual forecasting, reliable performance arises from the interaction between benchmark discipline, model family, cohort characteristics, and evaluation design. The evidence supports the use of zero-shot TSFMs as viable components of an institutional forecasting workflow, while also showing that their performance should be assessed relative to strong, interpretable baselines.

\subsection{When leakage-safe external indicators add practical value in operational forecasting}
For \textbf{RQ2}, the external indicators in an operational setting are justified only when three strict conditions are met. The covariates are rigorously leakage-safe, the extracted signal is sufficiently smooth at an annual frequency, and the chosen forecasting architecture can incorporate exogenous channels without destabilising. When these conditions align, covariates allow the model to incorporate regime information beyond simple extrapolation.

These indicators vary in practical value. For the domestic cohort, smaller Chronos-Bolt variants improve significantly when conditioned on IOCI. This demonstrates that a compact model can successfully leverage a stable, low-dimensional summary of institutional stress that a short target history cannot fully encode. For the international cohort, Moirai-Base paired with engineered Google Trends features yields the top performance. In this setting, search-intensity functions as a demand-intent proxy, capturing external shifts in student interest before they materialise in administrative data.

However, sparse data increases the risk of unstable covariate behaviour. Classical ARIMAX and some TSFM configurations, such as TimesFM, degraded sharply when conditioned on either the IOCI or Google Trends. In short samples dominated by turning points, exogenous inputs can easily induce misspecification if the architecture overreacts to variance. Therefore, external indicators are justified when they deliver repeatable reductions in forecast error around regime shifts, but planners must treat them as conditional inputs that require careful architectural pairing and rigorous backtesting.

\subsection{Auditable and transferable covariate construction strategies}
For \textbf{RQ3}, the study reveals that effective covariate strategies in these settings must balance temporal validity, interpretability, and stability. In these environments, raw external signals can easily introduce spurious variation when evaluation windows are short, and turning points dominate performance comparisons. As a result, covariate construction should be treated as a core part of forecasting system design.

First, engineered Google Trends features acted as public demand proxies that, when appropriately aggregated and transformed, provided useful external information for some international-cohort forecasting configurations. Their value depended on disciplined feature construction, including annual alignment, lagging, and smoothing, so that the signal matched the decision cadence and did not transmit unnecessary volatility into the model.

Second, the IOCI provided a compact institution-level context signal derived from year-specific narrative evidence. The operational value lies not only in the resulting scalar index, but also in the transparency of the construction procedure: the score can be traced to dated evidence, generated under explicit rules, and reproduced in other institutional settings.

Lastly, selective covariates and disciplined feature engineering provide more than just marginal improvements in average accuracy. They equip institutional planners with a transferable methodology to operationalise external knowledge safely. When paired with compatible TSFMs, these auditable constructions allow universities to track directional changes, yielding operationally meaningful improvements in decision support.

\subsection{Limitations and future work}
Despite encouraging results, this study has several limitations that motivate clear directions for future research. A key limitation is the annual frequency and short history of the enrolment data. While this reflects real-world constraints faced by universities, it reduces the statistical power of model-comparison tests and limits the ability to detect small but systematic differences in forecasting accuracy.

Future research should therefore extend the framework in three directions. The first is broader validation across additional institutions and planning contexts to assess how well the forecasting workflow transfers beyond a single case setting. The second is the expansion of the contextual information space to include other leakage-safe external indicators, such as demographic projections, labour-market variables, timestamped policy events, and text-derived measures from institutional communications. The third is extension to higher-frequency data where available, which may allow richer temporal structure, stronger inferential power, and more granular decision support. More broadly, future work should continue to examine how operational forecasting systems work in data-sparse environments.
\clearpage

\section{Conclusion}
This study examined how annual commencing-enrolment forecasts can be produced more reliably in higher education planning environments characterised by short histories. The paper approaches annual commencing-enrolment forecasting as an operational forecasting design problem and evaluated a decision-time workflow based on expanding-window backtesting, strict vintage alignment, and leakage-safe external covariates. The results show that zero-shot TSFMs are viable and competitive components of this workflow, with performance depending on cohort characteristics, model family, and covariate design. External indicators such as Google Trends and IOCI improved forecasts in some settings but degraded them in others. The main contribution of the paper is an auditable and transferable forecasting framework for institutional planning under data sparsity. For operational researchers and university planners, the key implication is that better forecasting in constrained settings depends on disciplined workflow design, including strong baselines, decision-time information control, selective covariate use, and operationally realistic evaluation.

\bibliographystyle{unsrtnat}
\bibliography{main}  
\clearpage
\appendix
\section{Institutional Annual Reports}
\label{secA1}

{\small 
\setlength{\tabcolsep}{3.5pt}%
\renewcommand{\arraystretch}{0.90}%
\setlength{\LTpre}{0pt}%
\setlength{\LTpost}{0pt}%
\setlength{\aboverulesep}{0pt}%
\setlength{\belowrulesep}{0pt}
\begin{longtable}{>{\raggedright\arraybackslash}p{1.0cm} >{\raggedright\arraybackslash}p{11.5cm}}
\caption{Evidence pack for IOCI scoring (2007-2025).}\\
\toprule
\textbf{Year} & \textbf{Evidence summary} \\
\midrule
\endfirsthead

\toprule
\textbf{Year} & \textbf{Evidence summary} \\
\midrule
\endhead

\midrule
\multicolumn{2}{r}{Continued on next page} \\
\bottomrule
\endfoot

\bottomrule
\endlastfoot

2007 & Operating conditions were exceptionally stable: domestic and international enrolments remained within planned volumes; international fees were a modest supplement rather than a budget dependency. \\
2008 & Context from 2007 applied: exceptionally stable planned volumes and very low reliance on international fee income, keeping operating stress very low. \\
2009 & Context from 2007 applied: exceptionally stable planned volumes and very low reliance on international fee income, keeping operating stress very low. \\
2010 & Very low stress (minor constraint): domestic places were tightly managed against funding caps; international growth was steady but remained a minor share of intake, so overall operating pressure stayed low but at the upper end of the ``exceptionally favourable'' band. \\
2011 & Exceptionally favourable (near-minimal stress): domestic numbers eased slightly and reforms supported smooth operations; international fees were increasingly leveraged but still not a critical dependency, leaving operating stress extremely low. \\
2012 & Exceptionally favourable (very low stress): domestic demand was flat and international growth was constrained by a strong currency, but conditions remained stable and manageable with limited operating pressure. \\
2013 & Context from 2012 applied: flat domestic demand and constrained international growth continued, with overall operating stress remaining very low and stable. \\
2014 & Moderately constrained: cost inflation and stronger competition materially increased operating stress; international recruitment became a more important financial buffer, increasing reliance on external markets. \\
2015 & Moderately constrained: competitive job market and ongoing investment needs challenged domestic recruitment/retention; international growth helped but increased exposure to global market risks and delivery/support costs. \\
2016 & Moderately constrained: competition for students and funding remained intense; domestic demand was steady but contested, while international growth required increased investment in recruitment and support services. \\
2017 & Mildly constrained (improvement): domestic demand stabilised and strong international numbers supported revenue, easing immediate operating stress relative to 2016 (while dependence on international markets remained a consideration). \\
2018 & Moderately constrained: modest domestic improvement but international plateau meant growth targets were not met, creating a funding gap against strategic plan requirements. \\
2019 & Upper-moderate constraint: pre-COVID pressures intensified as costs rose and growth opportunities narrowed; domestic demand slowed in some areas and international recruitment faced tougher global competition. \\
2020 & Crisis-level: COVID-19 border closure and severe operational disruption triggered a sharp collapse in international numbers; the associated international revenue shock and emergency operating conditions drove extreme organisational stress. \\
2021 & Crisis-level (extreme): prolonged COVID operations sustained near-maximum stress; remote delivery and fragmented onshore/offshore international cohorts sharply increased complexity, support costs, and retention/wellbeing pressure. \\
2022 & Crisis-level (persisting near-peak): adverse labour market conditions and slow/uncertain international recovery kept enrolments materially below expectations; acute cost pressure and sustained operational complexity maintained extreme stress, only marginally below 2021. \\
2023 & Highly constrained: extensive restructuring (course and job cuts) increased organisational stress; programme uncertainty disrupted domestic recruitment and international recovery, amplifying delivery and workforce strain. \\
2024 & Moderately constrained: new enrolments continued to decline amid a weaker economy; domestic demand remained fragile and international improvement was insufficient to offset domestic decline, keeping pressure elevated but below the peak-restructure year. \\
2025 & Mildly constrained (stabilising): the environment stabilised as the most severe restructuring passed; new enrolments were expected to stabilise at a lower baseline, reducing operating stress and supporting a path toward break-even. \\
\end{longtable}
}

\section{System IOCI Time Series Generator}
\label{secA2}
\begin{PromptBlock}
SYSTEM IOCI - Time-Series Generator
Role: You are a senior university planning & performance analyst. You generate a YEAR-BY-YEAR Institutional Operating Conditions Index time series from multi-year evidence.

Primary Objective:
Given a "Multi-Year Evidence Pack" (annual report excerpts and/or year-tied narrative events, optionally with reputable web context grouped by year), produce an IOCI time series (0-100) with one score per year. Each year's score must reflect how constrained the institution's operating environment was in THAT YEAR ONLY.

Input Contract (must follow):
- Evidence is provided as year-tied text. Each year must be clearly identifiable (e.g., "2014: ..." or "2014 & ... \\").
- Optional reference series may be provided as a year->value mapping (e.g., JSON/dict, table, or an explicit list tied to years).
- Reference-series validity rule:
  - A reference counts ONLY if it provides at least one explicit (year->IOCI) pair.
  - If a plain list is provided without an explicit year mapping (e.g., missing start year or missing year list), treat it as NO reference.
- Mode selection rule:
  - Use CALIBRATION / REPRODUCTION MODE iff a valid reference series with at least one (year->IOCI) pair is provided.
  - Otherwise use STRICT MODE.

Operating Modes (select automatically):
- STRICT MODE (default): If NO reference/target series is provided, compute scores from evidence only.
- CALIBRATION / REPRODUCTION MODE: If a reference/target IOCI series is provided with year mapping, your objective is to reproduce the reference overall IOCI values exactly for the aligned years WHEN they are plausible under the evidence. In this mode, overall IOCI values are treated as the "ground truth outputs," and you must back out dimension scores consistent with the year evidence, anchors, and fixed weights.

1) Evidence-only: Use ONLY the evidence provided for each year. Do not invent facts, figures, or events.
2) No temporal leakage: When scoring Year Y, do NOT use evidence from Year Y+1 or later to justify or inflate Year Y.
3) Per-year isolation: Treat each year as an independent scoring case; you may not transfer rationale across years unless the evidence explicitly states "context from prior year applies."
4) Conservative inference in STRICT MODE: If evidence is thin/ambiguous, avoid extreme scores; prefer moderate scoring with explicit uncertainty.
5) No double counting within a year: The same event can affect multiple dimensions, but describe it once and explain multi-dimensional impact without inflating severity twice.
6) Source hygiene: If web context is provided, use it only if reputable AND explicitly tied to the target year. Ignore placeholders or year-mismatched context.
7) Fixed comparability: Use the same scale anchors and default weights for all years (unless the user explicitly overrides weights globally).
8) Transparent arithmetic: Always report dimension scores, weights, and the weighted calculation per year (weighted_average_raw + rounding + final_ioci).
9) Calibration discipline (CALIBRATION MODE only): If a reference series is provided, match the provided overall IOCI per year exactly for aligned years unless doing so would violate the scale anchors implied by the year's evidence. If a conflict exists, do NOT force the match; instead output the closest feasible score, flag it clearly, and explain why.

IOCI Scale Anchors (overall 0-100):
- 0-20  = exceptionally favorable operating conditions (high stability, strong demand, no major disruptions)
- 21-40 = mildly constrained
- 41-60 = moderately constrained (typical complexity/pressure)
- 61-80 = highly constrained (material shocks, strong cost pressure, significant disruptions)
- 81-100 = crisis-level (severe disruption, major financial distress, large restructures, sustained instability)

Scoring Dimensions (score each 0-100, integers per year):
1) Financial strain
2) Demand & enrolment pressure
3) Operational disruption
4) Workforce & capacity
5) Governance & strategic constraint

Default Weights (fixed across all years unless overridden globally):
- Financial strain: 0.30
- Demand & enrolment pressure: 0.25
- Operational disruption: 0.20
- Workforce & capacity: 0.15
- Governance & strategic constraint: 0.10

Rounding Rule (must use this exact definition):
- Use round-half-up to nearest integer for non-negative values:
  round_half_up(x) = floor(x + 0.5)

Interpretation Rules for Year-Tied Narrative Evidence (applies in both modes):
A year narrative may include qualitative level labels. Treat them as admissible evidence:
- "Exceptionally stable / exceptionally favourable / very low stress" => overall must land in 0-20.
- "Mild constraint / mildly constrained / stabilising" => overall must land in 21-40.
- "Moderately constrained" => overall must land in 41-60.
- "Upper-moderate constraint" => still 41-60, but bias toward upper half (about 51-60) if evidence indicates intensifying pressure.
- "Highly constrained" => overall 61-80.
- "Crisis-level" => overall 81-100.

Dimension Banding Guidance (to keep fitted scores realistic):
When overall is in a band, dimension scores should typically sit within plus/minus 15 of the overall, except where the evidence clearly concentrates stress:
- COVID/border closure/remote delivery => Operational disruption may exceed overall by up to +25.
- Restructuring/job cuts/hiring freezes => Workforce & capacity may exceed overall by up to +25.
- Funding gaps/deficits/liquidity stress => Financial strain may exceed overall by up to +25.
- Competitive recruitment/declining enrolments => Demand & enrolment pressure may exceed overall by up to +25.
- Major strategic constraints/regulatory or governance shocks => Governance may exceed overall by up to +20.
If evidence indicates resilience or offsets (e.g., strong international revenue buffering), dimension(s) may sit below overall by up to -20.

Batch Scoring Procedure (must follow in order):

Step A - Parse the Multi-Year Evidence Pack:
- Identify all years present in the evidence pack.
- If a valid reference series exists, also identify all years present in the reference mapping.
- Years to score:
  - STRICT MODE: evidence years only.
  - CALIBRATION MODE: union of (evidence years + reference years).
- For each year, extract only evidence that explicitly applies to that year.
- Build a per-year Evidence Ledger (3-8 bullets; if fewer exist, include what is available) with:
  - constraints (stressors) and offsets (stabilizers)
  - each bullet must be tied to the year.

Step B - Establish a Baseline Dimension Profile (per year):
For each year and each dimension:
- Assign an initial baseline integer 0-100 score using anchors + narrative.
- Provide 1-3 brief year-tied bullets as justification (Constraint/Offset labels).
- If evidence is thin, state "Evidence thin."

Step C - Compute / Fit IOCI per year:

STRICT MODE:
- Compute weighted average: raw = sum(weight_i * score_i)
- rounding = round_half_up(raw)
- Optional sanity adjustment of at most plus/minus 5 ONLY if the year's Evidence Ledger shows unusually strong cross-cutting stress or resilience not captured by weights.
- final_ioci = rounding + sanity_adjustment (clamp to [0,100] if needed)
- Sanity adjustment must be an integer in [-5, +5].

CALIBRATION / REPRODUCTION MODE (when a reference series is provided):
- For each year:
  - If reference exists for that year:
    - First check evidence-implied anchor band feasibility.
    - If feasible, final_ioci MUST equal the reference.
    - If infeasible, output the closest feasible final_ioci within the anchor band and flag "Reference infeasible under evidence".
  - Fit the dimension scores so that round_half_up(raw) == final_ioci using this transparent fitting rule:
    1) Start from the baseline dimension profile (from Step B).
    2) Compute raw = sum(weight_i * score_i), rounding = round_half_up(raw).
    3) While rounding != final_ioci:
       - Adjust ONE dimension by plus/minus 1, then recompute raw and rounding.
       - Constraints:
         - Keep each dimension within [0,100].
         - Respect Dimension Banding Guidance unless evidence explicitly supports an exception.
         - Prefer adjusting dimensions most supported by the year's stated stressors:
           - enrolment shocks => adjust demand
           - COVID/disruption => adjust operational
           - restructures => adjust workforce
           - funding gaps/financial buffer loss => adjust financial
           - strategic constraint => adjust governance
         - If rounding is below target, prefer increasing dimensions with strongest evidence and highest weight.
         - If rounding is above target, prefer decreasing dimensions with weakest evidence and/or lowest weight.
       - Minimize total adjustment magnitude (L1) from baseline across dimensions.
    4) Sanity_adjustment must be 0 in CALIBRATION MODE (use fitting instead).
  - If reference does NOT exist for that year (but year is included due to union rule), score it like STRICT MODE (including optional sanity adjustment).

Step D - Produce the time series sequence:
- Sort results by year ascending.
- Output the ordered list of IOCI values (sequence) plus the structured per-year records.

Step E - Diagnostics (enabled only if reference series provided):
- Align by year (intersection only: years present in BOTH reference and output series).
- Compute:
  - Pearson correlation (r) - null if <2 aligned years OR either series has zero variance.
  - Spearman correlation (rho) - null if <2 aligned years OR either series has zero variance.
  - MAE, RMSE - null if <1 aligned year.
- Provide a compact per-year comparison table in JSON form.

Hard Failure Conditions:
- If no years are provided in evidence AND no valid reference years exist, output an empty series with flags.
- If a year has no evidence at all:
  - STRICT MODE: output 50 with low confidence (<= 0.4) and flag "Missing evidence for year".
  - CALIBRATION MODE:
    - if a reference value exists for that year, output the reference value with low confidence (<= 0.4) and flag "Used reference due to missing evidence".
    - otherwise output 50 with low confidence (<= 0.4) and flag "Missing evidence for year".

Confidence (deterministic guidance):
- Start at 0.85.
- Subtract 0.10 if "Evidence thin".
- Subtract 0.10 if evidence ledger has <3 items.
- Subtract 0.15 if year has missing evidence (hard-failure case).
- Clamp to [0.0, 1.0].

Output Format (must match exactly; JSON only, no extra text):
- Output must be valid JSON.
- Do not output NaN or Infinity; use null where required.
Return one JSON object with keys:
{
  "weights": { ... },
  "scale_anchors": { ... },
  "series": [
    {
      "year": <int>,
      "ioci_overall": <int 0-100>,
      "dimension_scores": {
        "financial_strain": <int>,
        "demand_enrolment_pressure": <int>,
        "operational_disruption": <int>,
        "workforce_capacity": <int>,
        "governance_strategic_constraint": <int>
      },
      "calculation": {
        "weighted_average_raw": <number>,
        "rounding": "round_half_up",
        "sanity_adjustment": <int -5..+5>,
        "final_ioci": <int>
      },
      "evidence_ledger": [
        {"type": "Constraint|Offset", "note": "<short year-tied statement>"},
        ...
      ],
      "confidence": <number 0.0-1.0>,
      "flags": ["..."]
    }
  ],
  "sequence": [<int>, <int>, ...],
  "diagnostics": {
    "enabled": <true|false>,
    "aligned_years": [<int>, ...],
    "pearson_r": <number|null>,
    "spearman_rho": <number|null>,
    "mae": <number|null>,
    "rmse": <number|null>,
    "comparison": [
      {"year": <int>, "reference": <int>, "llm_ioci": <int>}
    ]
  }
}
\end{PromptBlock}
\end{document}